\documentclass[journal]{IEEEtran}
\usepackage[T1]{fontenc}

\IEEEoverridecommandlockouts                          

\usepackage{tabularx}
\usepackage{booktabs}

\usepackage{amsmath} 
\usepackage{amssymb}  
\usepackage{xcolor}
\usepackage{caption}
\usepackage{graphicx}
\usepackage{mathtools}
\usepackage{caption}
\usepackage{floatrow}
\usepackage{tikz}

\usepackage{algorithmicx}
    \usepackage[ruled]{algorithm}
    \usepackage{algpseudocode}

\usepackage{pgfplots}
\usepackage{xcolor}
\usetikzlibrary{arrows.meta,positioning,shapes,intersections,patterns,calc,fit,decorations,decorations.markings} 
\usepackage{graphicx}

\usepackage{amsmath} 

\usepackage{enumerate}
\usepackage[all,tips]{xy}
\SelectTips{cm}{11}

\newtheorem{remark}{Remark}

\usepackage{bm}
\newcommand{\vect}[1]{\bm{#1}}
\newcommand{\matr}[1]{\bm{#1}}
\newcommand{\quat}[1]{\overline{\vect{#1}}}

\renewcommand{\S}{\matr{S}} 
\newtheorem{lemma}{Lemma}
\newtheorem{theorem}{Theorem}

\usepackage{bm}
\renewcommand{\S}{\matr{S}} 
\newcommand{\I}{\matr{I}}
\newcommand{\K}{\matr{K}}
\newcommand{\DQ}[1]{\vect{#1}}
\newcommand{\Q}{\DQ{Q}}

\newcommand{\PrincipalPart}{\mathcal{P}}
\newcommand{\DualPart}{\mathcal{D}}

\DeclareMathOperator{\sign}{sgn}

\newcommand{\R}{\mathbb{R}}

\newcommand{\N}{\mathbb{N}}

\newcommand{\D}{\mathcal{D}}

\newcommand{\Verts}[1]{{\left\Vert #1 \right\Vert}}

\newtheorem{assumption}{Assumption}

\newtheorem{rem}{Remark}
\usepackage[noabbrev]{cleveref} 
\crefname{rem}{Remark}{Remarks}
\crefname{exam}{Example}{Examples}
\crefname{assum}{Assumption}{Assumptions}
\crefname{prop}{Proposition}{Propositions}
\crefname{propy}{Property}{Properties}
\crefname{cor}{Corollary}{Corollaries}
\crefname{lem}{Lemma}{Lemmas}
\crefname{section}{Section}{Sections}
\crefname{thm}{Theorem}{Theorems}
\crefname{defn}{Definition}{Definitions}
\crefname{figure}{Fig.}{Fig.}
\Crefname{figure}{Figure}{Figures}
\crefname{equation}{}{}

\newcommand{\q}{\bm q}
\newcommand{\p}{\bm p}
\newcommand{\br}{\bm r}

\newcommand{\e}{\bm e}

\newcommand{\y}{\bm{y}}

\newcommand{\yd}{\y^{\dagger}}
\newcommand{\qd}{\overline{\delta \q}^{\dagger}}
\newcommand{\Qd}{\delta \Q^{\dagger}}
\usepackage{mathtools}
\usepackage{siunitx}

\begin{document}

\title{
Dual-quaternion learning control for autonomous vehicle trajectory tracking with safety guarantees}

\author{Omayra Yago Nieto$^{1}$, Alexandre Anahory Simoes$^{2}$, Juan I. Giribet$^{3}$, \emph{Senior Member, IEEE} and\\ Leonardo J. Colombo$^{4}$
\thanks{$^{1}$Omayra Yago Nieto is a PhD Student at Universidad Politécnica de Madrid, Spain. 
        {\tt\footnotesize omayra.yago.nieto@alumnos.upm.es}}
        \thanks{$^{2}$Alexandre Anahory Simoes is with the School of Science and Technology, IE University, Spain
        {\tt\footnotesize alexandre.anahory@ie.edu}}
        \thanks{$^{3}$Juan I. Giribet is with Universidad de San Andr\'es (UdeSA) and CONICET, Argentina.
        {\tt\footnotesize jgiribet@conicet.gov.ar}}
         \thanks{$^{4}$Leonardo J. Colombo is with the Centre for Automation and Robotics
(CSIC-UPM), Spain.
        {\tt\footnotesize leonardo.colombo@car.upm-csic.es}}
\thanks{J. Giribet was partially supported by  PICT-2019-2371 and PICT-2019-0373 projects from Agencia Nacional de Investigaciones Cient\'ificas y Tecnol\'ogicas, and UBACyT-0421BA project from the Universidad de Buenos Aires (UBA), Argentina. The authors acknowledge financial support from the Spanish Ministry of Science and Innovation, under grants  PID2022-137909NB-C21 funded by MCIN/AEI\-/10.13039\-/501100011033. The research leading to these results was supported in part by iRoboCity2030-CM, Robótica Inteligente para Ciudades Sostenibles (TEC-2024/TEC-62), funded by the Programas de Actividades I+D en Tecnologías en la Comunidad de Madrid.}}

\maketitle


\begin{abstract}
We propose a learning-based trajectory tracking controller for autonomous robotic platforms
whose motion can be described kinematically on $\mathrm{SE}(3)$.
The controller is formulated in the dual quaternion framework and operates at the velocity level,
assuming direct command of angular and linear velocities, as is standard in many aerial vehicles
and omnidirectional mobile robots.
Gaussian Process (GP) regression is integrated into a geometric feedback law to learn and
compensate online for unknown, state-dependent disturbances and modeling imperfections affecting
both attitude and position, while preserving the algebraic structure and coupling properties
inherent to rigid-body motion.

The proposed approach does not rely on explicit parametric models of the unknown effects,
making it well suited for robotic systems subject to sensor-induced disturbances,
unmodeled actuation couplings, and environmental uncertainties.
A Lyapunov-based analysis establishes probabilistic ultimate boundedness of the pose tracking error
under bounded GP uncertainty, providing formal stability guarantees for the learning-based controller.

Simulation results demonstrate accurate and smooth trajectory tracking in the presence of realistic,
localized disturbances, including correlated rotational and translational effects arising from
magnetometer perturbations.
These results illustrate the potential of combining geometric modeling and probabilistic learning
to achieve robust, data-efficient pose control for autonomous robotic systems.
\end{abstract}


\IEEEpeerreviewmaketitle

\section{Introduction}
\IEEEPARstart{T}{he} demand for unmanned aerial and underwater vehicles is rapidly increasing in many areas such as monitoring, mapping, agriculture, and delivery. The dynamics of these systems can often be described by rigid body motion, and most control approaches are based on feedback linearization~\cite{lee2009feedback} or backstepping methods~\cite{raffo2008backstepping}, often analyzed in terms of stability~\cite{frazzoli2000trajectory}. However, these classical methods rely on precise models of the vehicle dynamics and disturbances to guarantee stability and accurate tracking. Obtaining an exact model of such uncertainties through first-principles is often infeasible. In particular, aerodynamic or hydrodynamic effects, interactions with unstructured environments, and external perturbations introduce strong nonlinearities that are difficult to model. Increasing feedback gains to compensate for these uncertainties typically leads to noise amplification and actuator saturation. 

A promising alternative is offered by learning-based oracles, such as Neural Networks and Gaussian Processes (GPs), which can capture unknown dynamics directly from data. These data-driven models have achieved remarkable results in control, learning, and system identification. In the data-driven control paradigm, data collected from the system are used to infer and predict unknown dynamics in regions not covered by the training set. Unlike parametric models, data-driven approaches are highly flexible and capable of reproducing complex nonlinear behavior~\cite{hou2013model}. Among them, Gaussian Processes have recently gained attention for modeling dynamical systems due to their Bayesian formulation, bias-variance trade-off, and ability to provide uncertainty quantification~\cite{beckers2021introduction}. This last property is particularly appealing in control, where uncertainty estimates can be directly incorporated into stability and safety analyses. As a result, GP-based models have been successfully applied to predictive control~\cite{hewing2019cautious}, sliding mode control~\cite{lima2020sliding}, tracking of mechanical systems~\cite{beckers2019stable}, and backstepping control for underactuated vehicles~\cite{beckers2021online}. Their use to compensate model uncertainties in aerial and underwater vehicles has been further developed in~\cite{cao2017gaussian, beckers2021online, beckers2022safe, colombo2023learning}.

Rigid body attitude kinematics are typically represented by rotation matrices, unit quaternions, or local coordinates such as Euler angles. Due to the singularities associated with local coordinate charts, autonomous vehicle controllers often adopt the unit quaternion representation. More recently, dual quaternion formulations have emerged as a compact and singularity-free representation that jointly encodes both attitude and position in a single algebraic object~\cite{Wang2017, van2018attitude, wang2013, savino2020pose}. As unit dual quaternions form a Lie group isomorphic to $\mathrm{SE}(3)$, they provide a minimal and computationally efficient framework for representing and controlling rigid body motions~\cite{adorno2017cross, Lynch2017, kussaba2017hybrid}. This unified formulation has proven especially valuable in robotic control and navigation applications due to its geometric consistency and ease of interpolation.

In this work, we propose a learning-based trajectory tracking controller for
autonomous robotic platforms whose motion can be described kinematically on
$\mathrm{SE}(3)$.
The controller is formulated in the dual quaternion framework and operates at the
velocity level, assuming direct command of angular and linear velocities, as is
standard in many aerial vehicles and omnidirectional mobile robots.
Gaussian Processes (GPs) are integrated into a geometric feedback law to learn and
compensate unknown, state-dependent disturbances and modeling imperfections
affecting both attitude and translation.
By embedding GP-based disturbance estimates directly into the control loop, the
proposed method enables online adaptation using onboard data while preserving the
algebraic structure and coupling properties inherent to rigid-body motion.

From a practical robotics perspective, the proposed framework is motivated by
sensing- and actuation-related effects that are difficult to capture with
first-principles models at the kinematic level. As shownmin Section~IV,
localized disturbances affecting attitude estimation---such as magnetic anomalies
corrupting magnetometer measurements---can induce coupled rotational and
translational tracking errors, including altitude deviations in aerial vehicles.
Despite the absence of an explicit dynamic coupling model, the learning-based
controller is able to identify and compensate these correlated effects directly from data, resulting in improved pose tracking performance under realistic
operating conditions.

At the same time, the proposed method is supported by a rigorous theoretical
foundation.
Gaussian Processes provide not only data-driven regression of unknown kinematic
disturbances but also principled uncertainty quantification, which is explicitly
incorporated into the control design.
A Lyapunov-based analysis establishes probabilistic ultimate boundedness of the
pose tracking error under bounded GP uncertainty, yielding formal guarantees that
complement the empirical results observed in simulations.
This combination of geometric modeling, probabilistic learning, and stability
analysis enables a controller that is both practically effective for robotic
applications and theoretically well-founded.

To the best of our knowledge, this is the first work that combines online Gaussian
Process learning with dual quaternion--based feedback control at the kinematic
level to address uncertainty in pose tracking with probabilistic guarantees.
Previous studies have employed GPs for learning rigid-body dynamics
\cite{lang2015gaussian} and neural networks for predicting rigid-body motions in
dual quaternion form \cite{poppelbaum2022predicting}, but have not addressed the
integration of probabilistic learning within a geometric feedback control
framework with explicit stability guarantees, as presented in this paper.

The remainder of this paper is organized as follows. Section~\ref{sec:ps} introduces the problem formulation and preliminaries on dual quaternion kinematics and the nominal pose controller. Section~\ref{sec:lbc} presents the learning-based modeling framework using Gaussian Processes and the proposed tracking control law, together with the associated stability analysis. Section~\ref{sec:exp} reports simulation and experimental results demonstrating the performance and robustness of the proposed approach. Finally, conclusions and perspectives for extensions and applications are discussed in Section~V.


\section{Nominal controller in dual quaternions} \label{sec:ps}
\subsection{Nomenclature}

We begin by establishing the notation used throughout this section. 
The set $\mathbb{H}$ denotes the set of quaternions and $\mathcal{H}$ the set of dual quaternions. 
Unit quaternions and unit dual quaternions are denoted by $\mathbb{H}_u$ and $\mathcal{H}_u$, respectively.
Table~\ref{tbl:nomenclature} summarizes the symbols that will be used frequently.

\begin{table}[!htb]
\centering
\caption{Nomenclature for quaternions, dual quaternions and pose variables.}
\label{tbl:nomenclature}
\begin{tabularx}{\columnwidth}{c c X}
\toprule
\textbf{Symb.} & \textbf{Space} & \textbf{Description} \\
\midrule
$\vect{p}$ & $\mathbb{R}^3$ & Vehicle position in $\mathbb{R}^3$. \\
$\widetilde{\vect{p}}=(\vect{p},0)$ & $\mathbb{H}$ & Pure quaternion associated with $\vect{p}$. \\
\midrule
$\quat{q}=(\vect{q},q_0)$ & $\mathbb{H}$ & Quaternion with vector part $\vect{q}\in\mathbb{R}^3$ and scalar part $q_0\in\mathbb{R}$. \\
$\quat{q}^*$ & $\mathbb{H}$ & Quaternion conjugate of $\quat{q}$. \\
$\mathbb{H}_u$ &  & Set of unit quaternions. \\
\midrule
$\Q$ & $\mathcal{H}$ & Dual quaternion. \\
$\mathcal{H}_u$ &  & Set of unit dual quaternions. \\
$\PrincipalPart(\Q)$ & $\mathbb{H}$ & Principal (real) part of $\Q$. \\
$\DualPart(\Q)$ & $\mathbb{H}$ & Dual part of $\Q$. \\
\midrule
$\Q_d$ & $\mathcal{H}_u$ & Desired vehicle pose (dual quaternion). \\
$\delta \Q = \Q_d^* \circ \Q$ & $\mathcal{H}_u$ & Pose error (dual quaternion). \\
$\overline{\delta \q} = \q_d^*\circ \q$ & $\mathbb{H}_u$ & Attitude error quaternion. \\
$\widetilde{\delta \vect{p}}=(\delta\vect{p},0)$ & $\mathbb{H}$ & Position error as a pure quaternion. \\
\bottomrule
\end{tabularx}
\end{table}

\subsection{Generalities on dual quaternions}

Let $\vect{p}\in\mathbb{R}^3$ represent the vehicle position, and let
$a$ be a frame of reference. Then $\vect{p}^a$ denotes the vehicle
position expressed in frame $a$. 

Let $\mathbb{H}$ be the set of quaternions with the standard operations~\cite{Lynch2017, adorno2017robot}. The set $\mathbb{H}$ can be identified with $\mathbb{R}^4$ and its operations can be written in matrix form. In fact, every $\quat{q}\in\mathbb{H}$ can be decomposed in its vector component
$\vect{q}\in\mathbb{R}^3$ and real component $q_0\in\mathbb{R}$. Then,
given $\quat{p},\quat{q}\in\mathbb{H}$, with $\quat{p}=(\vect{p},p_0)$
and $\quat{q}=(\vect{q},q_0)$, the quaternion product, $\circ$, can be
written as
\begin{equation}\label{quat_prod}
    \quat{p}\circ\quat{q} =
        \begin{bmatrix}
            \S(\vect{p}) + \I p_0 & \vect{p} \\
                -\vect{p}^T      &    p_0
        \end{bmatrix}
        \begin{pmatrix}
            \vect{q} \\
            q_0
        \end{pmatrix},
\end{equation}
where $\I\in\mathbb{R}^{3\times 3}$ is the identity matrix.  The skew-symmetric matrix function $\S(\cdot)~:~\mathbb{R}^3\rightarrow\mathbb{R}^{3\times 3}$ is the matrix such that $\S(\vect{v})\vect{w}=\vect{v}\times \vect{w}$, for every $\vect{v},\vect{w}\in\mathbb{R}^3$, where $\times$ gives the vector product on $\mathbb{R}^3$. 

Given $\quat{\q}=(\q,q_0)\in \mathbb{H}$, the quaternion conjugate $\overline{\q}^{*}\in \mathbb{H}$ is defined as $\quat{\q}^{*}=(-\q,q_0)$. Notice that, given quaternions $\quat{\p},\quat{\q}$, then $(\quat{\p}\circ \quat{\q})^{*}=\quat{\q}^{*}\circ \quat{\p}^{*}$.

 The quaternion norm is given by $\Verts{\quat{\q}}^2 =\quat{\q}\circ \quat{\q}^* = \quat{\q}^* \circ \quat{\q}$. It is well known that the unit-norm quaternion represents rotations in the real $3$-sphere $\mathbb{S}^3$ in $\mathbb{R}^{4}$. The set of units quaternions will be denoted by $\mathbb{H}_{u}$ and it is an non-abelian Lie group under the quaternion multiplication.

 The rotation group $\mathrm{SO}(3)$ is equivalent to the real projective space $\mathbb{RP}^3$ 
of the antipodal point pairs on $\mathbb{S}^3$. 
Hence, the map $\mathcal{R} : \mathbb{H}_u \to \mathrm{SO}(3)$ defined by $\mathcal{R}(\quat q) = \{\pm\quat q\}$ is a $2$-to-$1$ homomorphism. In other words, $\quat{\q}\in \mathbb{H}$ and $-\quat{\q}\in \mathbb{H}$ represent the same rotation. 
The unit-norm quaternion and
corresponding rotation matrix are related by $\displaystyle{    \mathcal{R}(\quat{q})=(q_0^2-\vect{q}^T\vect{q}) \I + 2 \vect{q}\vect{q}^T +
    2 q_0 \S(\vect{q})}$. 


The Lie algebra of the unit quaternions $\mathbb{H}_u$ is
$\mathfrak{h}_u \cong \mathfrak{su}(2) \cong \mathfrak{so}(3) \simeq \mathbb{R}^3$,
where $\mathfrak{so}(3)$ denotes the Lie algebra of $\mathrm{SO}(3)$, i.e., the set of
$3\times 3$ skew-symmetric matrices. As a vector space, 
$\mathfrak{h}_u = \operatorname{Im}\mathbb{H} \cong \mathbb{R}^3$, the space of pure imaginary quaternions.

Every vector $\p\in \R^3$ can be identified with a quaternion {$\widetilde{\p}\in \mathbb{H}$ by $\widetilde{\p}=(\p,0)$. Suppose that $\quat{q}\in\mathbb{H}_u$ represents the rotation between the vehicle-frame $b$ and the inertial-frame
$i$, then $\widetilde{\p}^b$ and $\widetilde{\p}^i$ are related as $\widetilde{\p}^i = \quat{q} \circ \widetilde{\p}^b \circ \quat{q}^*$. Let $\widetilde{\vect{\omega}}=\vect{\omega}_{ib}^b (t)\in\mathfrak{h}_u\simeq\mathbb{R}^3$ be the
angular velocity of frame $b$ with respect to frame $i$, in frame
$b$. Then, the dynamics of $\quat{q}\in\mathbb{H}_u$, representing the
rotation from frame $b$ to frame $i$ is given by
\vspace{-2 mm} 
\begin{equation}\label{quat_dynamic}
    \dot{\quat{q}}  = \frac{1}{2} \quat{q} \circ \widetilde{\vect{\omega}}%
                    = \frac{1}{2}
                        \begin{bmatrix}
                            \S(\vect{q}) + \I q_0 \\
                            -\vect{q}^T
                        \end{bmatrix}\vect{\omega}.
\end{equation}


 Dual quaternions have been proved to be useful for vehicle position and attitude representation~\cite{adorno2017robot}. Let $\mathcal{H}$ be the set of dual quaternions, that is, the set 
\begin{equation}
    \mathcal{H}\triangleq \{ \Q = \quat{\q}_1+\varepsilon \quat{\q}_2 \ : \ \quat{\q}_1, \quat{\q}_2 \in \mathbb{H}, \ \varepsilon\neq 0, \ \varepsilon^2=0\},
\end{equation}
with $\varepsilon$ being an element having the following property: $\varepsilon\neq0$ and $\varepsilon^2=0$. We define the principal part as $\PrincipalPart(\Q)=\quat{\q}_1$ and the dual part as $\DualPart(\Q)=\quat{\q}_2$.
 

The sum, product, and conjugation of dual quaternions can be extended
from $\mathbb{H}$, taking into account that $\varepsilon^2=0$. 
The dual quaternion conjugate is given by
$\Q^* = \PrincipalPart(\Q)^* + \varepsilon \DualPart(\Q)^*$.  


In a similar way as unit-norm quaternions represents vehicle attitude, unit-norm dual quaternions are related to attitude and vehicle position, i.e., the pose of the vehicle. More precisely, let $\Q=\PrincipalPart(\Q) + \varepsilon \DualPart(\Q)$ a unit-norm dual quaternion satisfies $||\Q||^{2}=\Q\circ\Q^*=\Q^*\circ\Q=1$. The next result characterizes the unit-norm dual quaternions.

\begin{lemma}
The dual quaternion $\Q=\PrincipalPart(\Q) + \varepsilon \DualPart(\Q)$ has unit-norm if and only if its principal part $\PrincipalPart(\Q)$ is a unit-norm quaternion and the dual part can be written as $\DualPart(\Q)=\frac{1}{2}\widetilde{\p}\circ\PrincipalPart(\Q)$, where $\widetilde{\p}=(\vect{p},0)$ can be identified with $\vect{p}\in\mathbb{R}^{3}$.
\end{lemma}

\textit{Proof:} First, suppose that $\Q=\PrincipalPart(\Q) + \varepsilon\frac{1}{2}  \widetilde{\p}\circ\PrincipalPart(\Q)$, with $\widetilde{\p}=(\vect{p},0)$. Then $\Q\circ\Q^*=\PrincipalPart(\Q)\circ\PrincipalPart(\Q)^*+\varepsilon\frac{1}{2}  \left(\PrincipalPart(\Q)\circ\PrincipalPart(\Q)^*\circ\widetilde{\p}^* + \widetilde{\p}\circ\PrincipalPart(\Q)\circ\PrincipalPart(\Q)^*\right)=1$, because $\PrincipalPart{(\Q)}$ is a unit-norm quaternion and $\widetilde{\p}^*=-\widetilde{\p}$.

Suppose now that $\Q=\PrincipalPart(\Q) + \varepsilon \DualPart(\Q)$ is a unit-norm dual quaternion, it follows that $1=\Q\circ\Q^*=\PrincipalPart(\Q)\circ\PrincipalPart(\Q)^*+\varepsilon\frac{1}{2}  \left(\PrincipalPart(\Q)\circ\DualPart(\Q)^* + \DualPart(\Q)\circ\PrincipalPart(\Q)^*\right)$, then $\PrincipalPart(\Q)$ is a unit-norm quaternion and $\PrincipalPart(\Q)\circ\DualPart(\Q)^* + \DualPart(\Q)\circ\PrincipalPart(\Q)^*=0$. Let $\widetilde{\br}=\PrincipalPart(\Q)\circ\DualPart(\Q)^*$, then $\widetilde{\br}=-\widetilde{\br}^* $ i.e., $\widetilde{\br}=(\vect{r},0)$. Observe that,  $\DualPart(\Q)=-\widetilde{\br}\circ\PrincipalPart(\Q)$, then taking $\vect{p}=-2\vect{r}$, the unit-norm dual quaternion $\Q$ has the desired expression.
\hfill$\square$
\medskip 

In what follows, if $\quat{q}\in \mathbb{H}_u$ represents the attitude of the vehicle and $\widetilde{\p}\in\mathrm{Im}\,\mathbb{H}$, with $\mathrm{Im}\,\mathbb{H}$ the set of quaternions with real
part zero represents the vehicle position, the unit-norm dual quaternion will be expressed as $\Q = \quat{q} + \varepsilon \frac{1}{2} (\widetilde{\p}\circ\quat{q})$. 

The group of unit dual quaternions $\mathcal{H}_u$ is a Lie group that double-covers
the special Euclidean group $\mathrm{SE}(3)$, and thus provides a unified representation
of rigid body pose (position and attitude). Its Lie algebra $\mathfrak{k}_u$ can be
identified with purely imaginary dual quaternions,
\[
\mathfrak{k}_u = \Big\{\, Q = \tfrac{1}{2}(W + \varepsilon V)\;:\; W,V \in \mathrm{Im}\,\mathbb{H} \,\Big\},
\]
where $W$ encodes the angular velocity and $V$ the linear velocity.



Observe that given $\Q$, it is possible to recover vehicle attitude and 
position as follows: $\quat{q} = \PrincipalPart(\Q)$, and  $\widetilde{\p} = 2\DualPart(\Q) \circ \PrincipalPart{(\Q)}^*$. The time derivative of $\Q$ can be obtained with the
derivatives of the principal and dual parts. In fact,
\begin{equation}\label{quat-dual-dynamic}
\dot{\PrincipalPart}(\Q)  = \frac{1}{2}\PrincipalPart(\Q)\circ\widetilde{\vect{\omega}},\,\,
    \dot{\DualPart}(\Q) = \frac{1}{2}(\DualPart(\Q) \circ
       \widetilde{\vect{\omega}} + \widetilde{\vect{v}}\circ\PrincipalPart(\Q))
\end{equation}  where the notation $\widetilde{\vect{v}}=\dot{\widetilde{\vect{p}}}$ was introduced and the last equality
follows from equation~\eqref{quat_dynamic}. The time evolution of $\Q$, hence vehicle pose, is given
by the commanded angular velocity $\vect{\omega}$ (in vehicle frame) and
linear velocity $\vect{v}$ (in inertial frame). 

Let $\boldsymbol{\Omega}=\boldsymbol{\Omega}(\widetilde{\vect{\omega}},\widetilde{\vect{v}})$ be a dual quaternion, with principal and dual parts given by $\PrincipalPart(\boldsymbol{\Omega})=\widetilde{\vect{\omega}}$ and $\DualPart(\boldsymbol{\Omega})=\PrincipalPart(\Q)^*\circ\widetilde{\vect{v}}\circ\PrincipalPart(\Q)$, respectively, then
\begin{eqnarray}\label{vehic-dyn}
\dot{\Q}=\frac{1}{2}\Q\circ\bm{\Omega}(\widetilde{\vect{\omega}},\widetilde{\vect{v}}).
\end{eqnarray}

\begin{rem}
Throughout this paper, the angular velocity $\vect{\omega}=\vect{\omega}_{ib}^b$ is expressed in the body frame $b$,
whereas the linear velocity $\vect{v}=\dot{\vect{p}}$ is expressed in the inertial frame $i$.
Accordingly, the dual part of the twist dual quaternion is defined as
$\DualPart(\bm{\Omega})=\PrincipalPart(\Q)^*\circ \widetilde{\vect{v}}\circ \PrincipalPart(\Q)$,
so that $\bm{\Omega}$ is consistently represented in the body-related coordinates induced by $\Q$.
\end{rem}

\subsection{Dual quaternion error dynamics}

Let $\Q$ be the dual quaternion representing the current attitude and position of
vehicle and let $\Q_d$ be the desired dual quaternion, i.e., the desired vehicle pose. The dual quaternion error is  defined as 
$$\delta \Q= \Q_d^* \circ \Q = \overline{\delta \q} + \varepsilon\frac{1}{2}(\widetilde{\delta \p}^b\circ\overline{\delta \q}),$$ 
where $\overline{\delta \q}=\quat{\q}_d^*\circ\quat{\q}=(\delta \q,\delta
q_0)$ and $\widetilde{\delta \p} = \widetilde{\p}-\widetilde{\p}_d = (\delta \p,0)$, and
the term $\widetilde{\delta \p}^b = \quat{\q}_d^* \circ \widetilde{\delta \p} \circ
\quat{\q}_d = (\delta \p^b,0)$ can be interpreted as the position
error with respect to the desired vehicle frame. 
Observe that $\PrincipalPart(\delta \Q)=\overline{\delta \q}$ and
$\DualPart(\delta \Q)=\frac{1}{2}(\widetilde{\delta \p}^b\circ\overline{\delta
\q})$. 

We write $\overline{\delta \q}=(\delta \vect{q},\delta q_0)\in\mathbb{H}_u$ for the full attitude error quaternion,
where $\delta \vect{q}\in\mathbb{R}^3$ denotes its vector part and $\delta q_0\in\mathbb{R}$ its scalar part.


\vspace{.2cm}

\begin{lemma}\label{lemma_dyn}
Let $\widetilde{\vect{\omega}}   = (\vect{\omega}  , 0),\ \widetilde{\vect{v}}   = (\vect{v}  , 0)\in\mathbb{H}$, $\widetilde{\vect{\omega}}_d = (\vect{\omega}_d, 0),\ \widetilde{\vect{v}}_d = (\vect{v}_d, 0)\in\mathbb{H}$,
such that equation~\eqref{quat-dual-dynamic} or \eqref{vehic-dyn} are satisfied for $\Q=\quat{q}+\varepsilon\frac{1}{2}\widetilde{\p}\circ\quat{q}$ and $\Q_d=\quat{q}_d+\varepsilon\frac{1}{2}\widetilde{\p}_d\circ\quat{q}_d$,
respectively. Let the error dual quaternion $\delta \Q= \Q_d^* \circ \Q = \overline{\delta \q}+\varepsilon\frac{1}{2} \widetilde{\delta \p}^b \circ \overline{\delta \q}$. 
Given the dual quaternion $\delta\bf{\Omega}$, with $\PrincipalPart(\delta\boldsymbol{\Omega})=\widetilde{\vect{\omega}}-\overline{\delta \q}^*\circ\widetilde{\vect{\omega}}_d\circ\overline{\delta \q}$ and $\DualPart(\delta\boldsymbol{\Omega})=\overline{\delta \q}^{*} \circ \dot{\widetilde{\delta \p}}^b\circ\overline{\delta \q}$, the error $\delta \Q$ satisfies
\begin{align}\label{dual-quat-error}
    \dot{\delta\Q}=\frac{1}{2}\delta\Q\circ\delta\boldsymbol{\Omega}.
\end{align}

\end{lemma}

\textit{Proof:} 
By equation~\eqref{quat_dynamic} it follows that
\begin{align*}
    \dot{\PrincipalPart}(\delta \Q) &= \dot{\overline{\delta \q}}%
        = \dot{\quat{q}}_d^* \circ {\quat{q}} + \quat{q}_d^* \circ \dot{\quat{q}}%
        = \frac{1}{2}\left(%
                - \widetilde{\vect{\omega}}_d \circ \overline{\delta \q}
                + \overline{\delta \q} \circ \widetilde{\vect{\omega}} 
            \right)\\
            &=\frac{1}{2}\overline{\delta \q}\circ\left(\widetilde{\vect{\omega}}-\overline{\delta \q}^* \circ\widetilde{\vect{\omega}}_d\circ\overline{\delta \q}\right)=\frac{1}{2}\PrincipalPart(\delta\Q)\circ\PrincipalPart(\delta\boldsymbol{\Omega}).\\
        \dot{\DualPart}(\delta \Q) &= \frac{1}{2}\left(\widetilde{\delta \p}^b\circ\dot{\overline{\delta \q}}+\dot{\widetilde{\delta \p}}^b\circ\overline{\delta \q}\right)=\\
        &=\frac{1}{2}\left(
            \frac{1}{2}\widetilde{\delta \p}^b\circ\overline{\delta \q}\circ\PrincipalPart(\delta\boldsymbol{\Omega})+
            \overline{\delta \q}\circ\overline{\delta \q}^*\circ\dot{\widetilde{\delta \p}}^b\circ\overline{\delta \q}\right)\\
&=\frac{1}{2}\left(\DualPart(\delta \Q)\circ\PrincipalPart(\delta\boldsymbol{\Omega})+\PrincipalPart(\delta \Q)\circ\DualPart(\delta\Omega)\right).
\end{align*}
Then, observe that 
\[\dot{\PrincipalPart}(\delta\Q)+\varepsilon\dot{\DualPart}(\delta\Q)=\frac{1}{2}(\PrincipalPart(\delta\Q)+\varepsilon\DualPart(\delta\Q))\circ(\PrincipalPart(\delta\boldsymbol{\Omega})+\varepsilon\DualPart(\delta\boldsymbol{\Omega})),\]
and equation \eqref{dual-quat-error} follows.
\hfill$\square$

\begin{rem}
When the error velocity is given in inertial frame, an alternative expression for $\DualPart(\delta\boldsymbol{\Omega})$ can be obtained in Lemma \ref{lemma_dyn}. 

First, notice that $\dot{\quat{q}}_d^*\circ\widetilde{\delta \p}\circ\quat{q}_d+
{\quat{q}_d}^*\circ\widetilde{\delta \p}\circ\dot{\quat{q}}_d=(S(\delta \p^b)\vect{\omega}_d,0)$, then
{\begin{align*}
\dot{\widetilde{\delta \p}}^b&=\dot{\quat{q}}_d^*\circ\widetilde{\delta \p}\circ\quat{q}_d+
{\quat{q}_d}^*\circ\widetilde{\delta \p}\circ\dot{\quat{q}}_d+
{\quat{q}_d}^*\circ\dot{\widetilde{\delta \p}}\circ{\quat{q}_d},\\
&=(S(\delta \p^b)\vect{\omega}_d,0)+{\quat{q}_d}^*\circ\dot{\widetilde{\delta \p}}\circ{\quat{q}_d},
\end{align*}}
\noindent and it follows that 
$$\DualPart(\delta\boldsymbol{\Omega})=\overline{\delta \q}^*\circ (S(\delta \p^b)\vect{\omega}_d,0)\circ\overline{\delta \q} + \quat{q}^* \circ \widetilde{\delta \vect{v}} \circ \quat{q},$$ 
\noindent where $\widehat{\delta \vect{v}}=\dot{\widehat{\delta \p}}$ is the error velocity expressed in inertial frame.

Finally, we observe that
    \begin{equation}
        \dot{\overline{\delta \q}} = \frac{1}{2}\overline{\delta \q}\circ {\widetilde{\vect{\omega}}}- \frac{1}{2}\overline{\delta \q}\circ\overline{\delta \q}^*\circ{\widetilde{\vect{\omega}}_d}\circ\overline{\delta \q}. \label{eq:din_deltaq}
    \end{equation}\end{rem}

\subsection{Nominal control design}

In this Section we design a kinematic controller for the vehicle pose represented by the dual quaternion $\Q$, whose evolution is given by \eqref{vehic-dyn}. Given the desired pose of the vehicle $\Q_d$, the following theorem states the angular velocity and linear velocity
that must be applied to the vehicle in order to achieve the desired
attitude and position, i.e., $\Q\rightarrow\Q_d$.

\vspace{.2cm}

\begin{theorem}\label{theorem_1}
Let $\vect{\omega}_d,
\vect{v}_d$ be the desired angular and linear velocities, and the desired dual quaternion $\Q_d$ satisfying equation
~\eqref{vehic-dyn}. Let $\K_{\vect{\omega}},\K_{\vect{v}}:\mathbb{R}\rightarrow \mathbb{R}^{3\times 3}$ be uniformly bounded and uniformly positive definite matrix functions,
i.e., there exist constants $0<\alpha\leq \beta$ such that
\[
\alpha \I \preceq \K_{\vect{\omega}}(\rho) \preceq \beta \I,
\qquad
\alpha \I \preceq \K_{\vect{v}}(\rho) \preceq \beta \I,
\quad \forall\,\rho\in\mathbb{R},
\]
 and 
suppose that the dual quaternion $\Q$ is
given by equation~\eqref{vehic-dyn} with:
\begin{align}
   \widetilde{\vect{\omega}}&= (\vect{\omega}, 0)
        = \overline{\delta \q}^*\circ\widetilde{\vect{\omega}}_d\circ\overline{\delta \q} - ( \sign(\delta q_0) \K_{\vect{\omega}}(\rho)
            \delta \q, 0), \label{omegabar}\\
  \widetilde{\vect{v}}&=  (\vect{v}, 0)
        = \left(\widetilde{\vect{v}}_d - \, \K_{\vect{v}}(\rho)\delta \p
            , 0\right),\label{vbar}
\end{align}
\noindent where $\delta \Q = \Q_d^* \circ \Q=\overline{\delta \q}+\varepsilon\frac{1}{2} \widetilde{\delta \p}^b \circ \overline{\delta \q}$ and $\sign(x)$ is the sign function such that $\sign(0) = 1$. Then, $(\delta \q, \delta \p) \rightarrow (\textbf{0}, \textbf{0})$.


\end{theorem}

\begin{remark}The control objective is to track a desired pose $\Q_d$ and velocities $(\vect{\omega}_d, \vect{v}_d)$, i.e.,
to ensure $\delta \Q \to 1$ or, equivalently, $(\delta \q,\delta \p)\to(\bf{0},\bf{0})$.

Due to the double cover $\mathbb{H}_u \to \mathrm{SO}(3)$, the quaternions $\overline{\delta \q}$ and $-\overline{\delta \q}$ represent the same rotation.
The term $\sign(\delta q_0)$ enforces a shortest-rotation representation and prevents unwinding phenomena \cite{bhat2000topological}.\end{remark}

\vspace{.1cm}

\textit{Proof:} First, observe that $\DualPart(\delta\Q)^*\circ\DualPart(\delta\Q)=\frac{1}{4}\|\widetilde{\delta \p}\|^2$ and define the positive Lyapunov function $$ V(\delta\Q) =\ \delta \q^T\delta \q
       + 2\DualPart(\delta\Q)^*\circ\DualPart(\delta\Q).
        $$ The function $V$
satisfies $V = 0$ if and only if $\delta \q =  \vect{0}$, $|\delta q_0|=1$ and $\DualPart(\delta\Q)=(\vect{0},0)$. 

Taking the time derivative of $V$, by Lemma~\ref{lemma_dyn} it follows that
\begin{align*}
    \dot{V}
        =&\ \delta \q^T(\S(\delta \q) +
            \I\delta q_0)\PrincipalPart(\delta\boldsymbol{\Omega}) + \delta \p^{T} \dot{\delta \p},
\end{align*}
Because  $\dot{\vect{\delta p}}=\vect{v}-\vect{v}_d=-\K_{v}(\rho)\delta \p$, $\vect{x}^T \S(\vect{x}) = \vect{0}$ for every
$\vect{x}\in\mathbb{R}^3$, and $\PrincipalPart(\delta\boldsymbol{\Omega})=-\sign(\delta q_0) \K_{\omega}(\rho)\delta \q$,
it follows that
$$\dot{V} = - |\delta q_0| \delta \q^T \K_{\omega}(\rho)\delta \q - \delta \p^T \K_{v}(\rho) \delta \p.$$
Since $\K_v, \K_\omega$ are uniformly bounded, there exists  $\alpha>0$, such that $\K_v(\rho)\geq\alpha\I$ and $\K_\omega(\rho)\geq\alpha\I$, for every $\rho$, then $$\dot{V} =-\alpha\left(|\delta q_0|\|\delta \q\|^2+\|\delta \p\|^2\right)\leq 0,$$
\noindent for every $\delta \p \neq \vect{0}$ and $\delta \q \neq \vect{0}$. 

When $\delta q_{0}=0$ and $\delta \vect{p}=0$, we have $\dot{V}=0$. Substituting this and $\widetilde{\vect{\omega}}$ in \eqref{eq:din_deltaq} and also using that $\sign(0)=1$, we get
$$\dot{\overline{\delta \q}} = (-\K_{\vect{\omega}}(\rho)
            \delta \q, 0)$$
which implies $\dot{\delta \vect{q}}=-\K_{\vect{\omega}}(\rho)\delta \q$. In addition, we also have that $\|\delta \vect{q}\|=1$. Differentiating it over the solutions of the previous ODE, we obtain that $0=-\delta \vect{q}^{T}\K_{\vect{\omega}}(\rho)\delta \q$. Since $\K_{\vect{\omega}}(\rho)> 0$ this is only possible if $\delta \vect{q}=0$. Then, by LaSalle's Invariance principle, we deduce that the system's trajectories converge asymptotically to $\delta \vect{q} =  \vect{0}$ and $\delta \vect{p} =  \vect{0}$. \hfill$\square$

\subsection{Disturbance and measurement error model}

The underlying assumption in Theorem \ref{theorem_1} is that it is possible to measure the pose (i.e., the position and the attitude) $\quat{q}$ of the vehicle. Then it is possible to feedback the signal error $\delta \vect{q}$ in equation \eqref{omegabar}. 

Now, suppose that the attitude measurement $\quat{q}$ is corrupted by an unknown attitude error $\quat{q}_\rho$. In this case, the available signal is no longer $\overline{\delta \q}=\quat{q}_d^*\circ\quat{q}$, but
\[
\overline{\delta \q}_\rho := \quat{q}_d^*\circ\quat{q}\circ\quat{q}_\rho = \overline{\delta \q}\circ\quat{q}_\rho.
\]
Although the same could be done for position, here we explicitly consider the effect of both attitude and position errors in the complete pose, represented by the dual quaternion
\[
\Q_\rho := \quat{q}_\rho + \varepsilon\,\tfrac{1}{2}(\widetilde{\vect{p}}_\rho\circ\quat{q}_\rho),
\]where $\widetilde{\vect{p}}_\rho$ models possible translational errors (or biases) in the position measurement. 
Accordingly, the measured pose is
\[
\Q_{\text{meas}} = \Q\circ\Q_\rho,
\]
and the pose error used in feedback becomes
\[
\delta\Q_\rho := \Q_d^*\circ\Q_{\text{meas}} 
= \Q_d^*\circ\Q\circ\Q_\rho 
= \delta\Q\circ\Q_\rho.
\]

The explicit relation between $\delta\Q$ and $\delta\Q_\rho$, including their
principal and dual parts, is given in the Appendix.

If we control the system using the corrupted measurement $\Q\circ\Q_\rho$, then equations \eqref{omegabar} and \eqref{vbar} are replaced by:
\begin{align}
    &\widetilde{\vect{\omega}}_\rho = \widetilde{\vect{\omega}} + \vect{\eta_\omega}(\overline{\delta \q}_\rho), \label{omegabarerror}\\
    &\widetilde{\vect{v}}_\rho = \widetilde{\vect{v}} + \vect{\eta_{v}}(\delta \Q_\rho), \label{verror}
\end{align}

\noindent where $\vect{\eta_\omega}:\mathbb{H}_u\to\mathfrak{h}_u$ and $\vect{\eta_v}:\mathcal{H}_u\to\mathrm{Im}\,\mathbb{H}$ are state-dependent unknown functions that account for unmodeled disturbances and sensor errors due to $\quat{q}_\rho$ and $\Q_\rho$, respectively. 
We assume that the unknown terms can be decomposed as
\[
\vect{\eta_\omega}(\overline{\delta \q}_\rho)=\vect{\rho_\omega}(\overline{\delta \q}_\rho)+\nu_\omega,
\qquad
\vect{\eta_v}(\delta\Q_\rho)=\vect{\rho_v}(\delta\Q_\rho)+\nu_v,
\]
where $\nu_\omega$ and $\nu_v$ are zero-mean Gaussian disturbances with known covariance,
e.g., $\nu_\omega\sim\mathcal{N}(0,\Sigma_\omega)$ and $\nu_v\sim\mathcal{N}(0,\Sigma_v)$.
This corresponds to noisy training data for the GP models, with known noise statistics.

That is, the samples used to train the GP are noisy, and both the mean and variance of the noise are assumed known. The time dependency of the states and disturbances is omitted for simplicity of notation, i.e. $\vect{\eta_\omega}(\overline{\delta \q}_\rho):=\vect{\eta_\omega}(\overline{\delta \q}_\rho(t),t)$ and $\vect{\eta_v}(\delta\Q_\rho):=\vect{\eta_v}(\delta \Q_\rho(t),t)$.


Potential sources of $\vect{\tilde\rho_\omega}$ depend on the sensing instrument; 
for instance, they may model magnetic field disturbances when magnetometers are used as attitude sensors. 
Similarly, $\vect{\tilde\rho_v}$ may arise from unmodeled effects in the translational dynamics, 
such as aerodynamic drag, ground-effect interactions, or biases in GPS and visual odometry sensors. 
From now on, and to simplify notation, we omit the dependency on $\overline{\delta \q}_\rho$ and $\delta\Q_\rho$, 
and simply write $\overline{\delta \q}$ and $\delta\Q$.

\section{Learning-based trajectory tracking control based on Gaussian processes}
\label{sec:lbc}

As established in the vehicle dynamics \eqref{omegabarerror}–\eqref{verror}, 
the system is affected by unknown functions 
$\vect{\rho}_{\vect{\omega}}$ and $\vect{\rho}_{\vect{v}}$ 
that represent unmodeled or uncertain dynamics. 
The goal of this section is to design a controller that compensates these unknown effects 
by leveraging Gaussian Process (GP) regression to estimate them online.

The proposed strategy integrates the GP model within the control loop, 
updating the nominal controller with data-driven estimates of the disturbances. 
In contrast to continuous data–acquisition schemes, 
the vehicle does not need to record data at all times. 
To reduce computational load, data collection can be selectively activated, 
resulting in batches of measurements that are later used to update the GP model. 
This enables both online and hybrid offline/online learning phases. See Figure \ref{fig:control_graph}.

The GP–based predictions are embedded in the control law 
to correct for modeling errors and external perturbations that affect the vehicle dynamics. 
In this way, the controller continuously refines its internal model 
based solely on its own observed data, without requiring any external supervision 
or prior knowledge of the disturbance characteristics.

\begin{figure}[H]
    \centering
   \resizebox{0.99\textwidth}{!}{
\begin{tikzpicture}[
  block/.style={draw, minimum width=2.3cm, minimum height=1.2cm, align=center},
  sum/.style={draw, circle, minimum size=6mm, inner sep=0pt},
  arrow/.style={->, >=Stealth},
  dashedbox/.style={draw, dashed, inner sep=6pt, rounded corners=2pt, inner xsep=20pt},
  node distance=2cm and 2cm
]

\node[sum] (sum) at (0,0) {};
\node at ([xshift=-0.5cm,yshift=-0.3cm]sum.center) {\small $-$}; 

\draw[arrow] (-1.5,0) -- node[above] {$\Q_d,\vect{w}_d, \vect{v}_d$} (sum.west);       
\draw[arrow] (0,-1.5) -- node[pos=1.4, right, xshift=15pt] {$\delta \Q$} (sum.south);       

\node[block, right=of sum, font=\normalsize] (feedback) {Feedback};
\node[block, right=of feedback, font=\normalsize] (knowndyn) {Known\\dynamics};
\node[block, below=1.6cm of knowndyn, font=\normalsize] (unknowndyn) {Unknown\\dynamics};
\node[coordinate, right=of knowndyn] (output) {};
\node[below right=1.5cm and 1.1cm of output] {$[\Q,\dot{\Q}]$}; 


\coordinate (blacklinebase) at ($(knowndyn.east)+(2,0)$);

\draw[line width=3pt] ($(blacklinebase)+(0,1)$) -- ($(blacklinebase)+(0,-1)$);

\draw[] ($(blacklinebase)+(0,-0.6)$) -- ++(1.17,0);

\coordinate (feedbackReturnStart) at ($(blacklinebase)+(1.17,-0.6)$);
\draw[arrow] (feedbackReturnStart) -- ++(0,-3.2) -| (sum.south);

\draw[arrow] (sum) -- (feedback);
\draw[arrow] (feedback) -- node[above, xshift=-8pt] {$\widetilde{\vect{\omega}}_{c}$, $\widetilde{\vect{v}}_{c}$} (knowndyn);
\draw (knowndyn.east) -- (output.west);
\draw[arrow] (unknowndyn.north) -- node[right] {$\vect{\rho_\omega}(\overline{\delta \q})$, $\vect{\rho}_v(\delta \Q)$} (knowndyn.south);

\draw[arrow] (knowndyn.east) ++(0.3,0) |- (unknowndyn.east);



\node[dashedbox, fit={(knowndyn) (unknowndyn)}] (vehiclebox) {};

\node[anchor=east, text=black,font=\normalsize] at ($(vehiclebox.west)+(-0.2,0)$) {Vehicle};


\node[block, draw=violet, text=violet, above=of feedback] (vehiclemodel) {Vehicle\\model};
\draw[arrow, draw=violet] (vehiclemodel.south) -- (feedback.north);

\node[block, draw=violet, text=violet, font=\normalsize, above=1.5 of knowndyn] (gpmodel) {Dataset};
\node[block, draw=violet, text=violet, font=\normalsize, above=1.2 of gpmodel] (dataset) {GP\\Model};


\draw[draw=violet] 
  (gpmodel.south) -- ++(0,-0.4) coordinate (gpcontact);

\draw[draw=violet, fill=white] 
  (gpcontact) circle (2pt);

  \draw[draw=violet] 
  (dataset.south) -- ++(0,-0.4) coordinate (datacontact);

\draw[draw=violet, fill=white] 
  (datacontact) circle (2pt);

\draw[draw=violet] 
  ($(knowndyn.north) + (0,0.2)$) -- ++(0,0.4) coordinate (datacontact);

\draw[draw=violet, fill=white] 
  (datacontact) circle (2pt);

\draw[draw=violet] 
  (gpmodel.north) -- ++(0,0.4) coordinate (gpupcontact);

\draw[draw=violet, fill=white] 
  (gpupcontact) circle (2pt);
\draw[draw=violet, line width=0.5pt] 
  ($(gpupcontact)+(0.1,0)+(135:2pt)$) -- ++(60:8pt);

\draw[draw=violet, line width=0.5pt] 
  ($(datacontact)+(0.1,0)+(150:2pt)$) -- ++(60:9pt);

  \node[anchor=east, text=violet, font=\footnotesize, align=right] 
  at ($(gpupcontact)+(-0.1,0.2)$) {GP\\update};
  \node[anchor=east, text=violet, font=\footnotesize, align=right] 
  at ($(datacontact)+(-0.1,0.2)$) {Collection};

\draw[->, draw=violet, >=Stealth]
  ($(blacklinebase)+(1.2,0.95)$) -- 
  ++(0,4.2) coordinate (plineup) -- 
  ++(-3.2,0);

\node[text=violet] at ($(plineup)+(0.2,-1.3)$) {$\Q$};

\draw[draw=violet] 
  ($(blacklinebase)+(0.05,0.5)$) -- ++(1.15,0) -- ++(0,1.2);

  \draw[draw=violet]
  (dataset.west) -- ++(-5,0) coordinate (fhpoint)
  node[pos=0.8, above, text=violet] {$\widehat{\vect{\rho}}_{\vect{\omega},n}$, $\widehat{\vect{\rho}}_{\vect{v},n}$}
  -- ++(0,-1);

  \draw[->, draw=violet,  >=Stealth]
  ($(fhpoint)+(0,-1)$) -- ++(3.5,-3);
\end{tikzpicture}}
    \caption{Block diagram of the proposed control law}
    \label{fig:control_graph}
\end{figure}

Next, we present the learning and control framework in detail.

\subsection{Learning with Gaussian Processes}


Gaussian Processes (GPs) are stochastic processes fully specified by a mean function and a kernel function. 
To compensate the unknown dynamics in \eqref{omegabarerror}–\eqref{verror}, two independent GPs are defined to model 
$\vect{\rho_\omega}$ and $\vect{\rho_v}$, respectively. 
Their definitions are summarized in Table~\ref{tab:gp_def}.

\begin{table}[h!]
\centering
\caption{GP models used to compensate the unknown dynamics.}
\label{tab:gp_def}
\renewcommand{\arraystretch}{1.15}
\begin{tabular}{ll}
\hline
\textbf{Unknown function} & $\bm{\rho}_{\vect{\omega}}: \mathbb{H}_u \to \mathbb{R}^3$\\
\textbf{Inputs} & $x=\overline{\delta\q}\in\mathbb{H}_u\subset\mathbb{R}^4$\\
\textbf{GPs (per component)} & $\{f_{GP,i}^{(\vect{\omega})}\}_{i=1}^3$\\
\textbf{Kernel} & $k^{(\vect{\omega})}:\mathbb{R}^4\times\mathbb{R}^4\to\mathbb{R}$\\
\hline
\textbf{Unknown function} & $\bm{\rho}_{\vect{v}}: \mathcal{H}_u \to \mathbb{R}^3$\\
\textbf{Inputs} & $x=\delta\Q\in\mathcal{H}_u\subset\mathbb{R}^8$\\
\textbf{GPs (per component)} & $\{f_{GP,i}^{(\vect{v})}\}_{i=1}^3$\\
\textbf{Kernel} & $k^{(\vect{v})}:\mathbb{R}^8\times\mathbb{R}^8\to\mathbb{R}$\\
\hline
\end{tabular}
\end{table}

The unknown functions $\bm{\rho}_{\vect{\omega}}$ and $\bm{\rho}_{\vect{v}}$ are $\mathbb{R}^3$-valued.
In practice, we model each output component with an independent scalar GP that shares the same kernel,
i.e., for $(\cdot)\in\{\vect{\omega},\vect{v}\}$ and $i\in\{1,2,3\}$ we use
\[
f_{GP,i}^{(\cdot)}(x)\sim \mathcal{GP}\!\big(m_{GP,i}^{(\cdot)}(x),\,k^{(\cdot)}(x,x')\big),
\]
and stack the three posterior means to form the vector prediction.
This corresponds to the use of the columns $Y^{(\cdot)}_{:,i}$ in \eqref{eq:gp_mean_pred}--\eqref{eq:gp_var_pred}.

Each GP satisfies, for any $x,x'$ in its corresponding domain:
\begin{equation}
    f_{GP}^{(\cdot)}(x)\sim \mathcal{GP}\big(m_{GP}^{(\cdot)}(x),\,k^{(\cdot)}(x,x')\big),
\end{equation}
with
\begin{align}
    m_{GP}^{(\cdot)}(x) &= \mathbb{E}[f_{GP}^{(\cdot)}(x)], \label{eq:gp_mean}\\
    k^{(\cdot)}(x,x') &= \mathbb{E}\big[(f_{GP}^{(\cdot)}(x)-m_{GP}^{(\cdot)}(x))
    (f_{GP}^{(\cdot)}(x')-m_{GP}^{(\cdot)}(x'))\big]. \label{eq:gp_kernel}
\end{align}

Training datasets are constructed from $N(n)$ samples as
\begin{equation}
\mathcal{D}_{n(t)}^{(\cdot)}=\{x^{\{i\}},y_{(\cdot)}^{\{i\}}\}_{i=1}^{N(n)},
\,\,\,
y_{(\cdot)}^{\{i\}}=\bm{\rho}_{(\cdot)}(x^{\{i\}})+\nu_{(\cdot)}^{\{i\}},
\label{eq:gp_dataset}
\end{equation}
where $\nu_{(\cdot)}^{\{i\}}\sim\mathcal{N}(0,\sigma_{\nu,(\cdot)}^2 I_3)$ are i.i.d.\ measurement noises.
The inputs are $x=
\begin{cases}
\overline{\delta\q}, & \text{for } (\cdot)=\vect{\omega},\\
\delta\Q, & \text{for } (\cdot)=\vect{v}.
\end{cases}$

The datasets $\mathcal{D}_{n(t)}^{(\vect{\omega})}$ and $\mathcal{D}_{n(t)}^{(\vect{v})}$, with 
$n:\R_{\geq 0}\to\N$, can evolve over time $t$. 
At a given instant $t_1\in\R_{\geq 0}$, the dataset $\mathcal{D}_{n(t_1)}^{(\cdot)}$ with 
$N(n(t_1))$ training points exists. 
This construction allows the accumulation of training data over time, i.e., the number of training points 
$N(n)$ in $\mathcal{D}_{n}^{(\vect{\omega})}$ and $\mathcal{D}_{n}^{(\vect{v})}$ can be monotonically increasing, 
while also enabling a ``forgetting'' mechanism to keep $N(n)$ constant if desired.

For notational simplicity, we define $\mathcal{D}_{n}^{(\vect{\omega})}$ and 
$\mathcal{D}_{n}^{(\vect{v})}$ as the training datasets for the time interval 
$t\in[t_{n},t_{n+1})$, each containing $N(n)$ training points, where 
$0=t_{0}<t_{1}<t_{2}<\dots$. 
For instance, $\mathcal{D}_{n}^{(\vect{\omega})}$ may include all accumulated recorded pairs 
$\{\overline{\delta \q}^{\{i\}},\widehat{\y}_{(\vect{\omega})}^{\{i\}}\}$ up to time $t_{n}$, i.e., $\mathcal{D}_{0}^{(\vect{\omega})}\subset\dots \subset \mathcal{D}_{n}^{(\vect{\omega})}$,while $\mathcal{D}_{n}^{(\vect{v})}$ may contain all accumulated pairs 
$\{\delta \Q^{\{i\}},\widehat{\y}_{(\vect{v})}^{\{i\}}\}$ up to $t_{n}$, i.e., $\mathcal{D}_{0}^{(\vect{v})}\subset\dots \subset \mathcal{D}_{n}^{(\vect{v})}$.

Alternatively, the datasets may be disjoint if each $\mathcal{D}_{n}^{(\cdot)}$ consists only of 
newly recorded data.

The time-dependent GP estimates are denoted by 
$\widehat{\bm{\rho}}_{\vect{\omega},n}$ and $\widehat{\bm{\rho}}_{\vect{v},n}$ 
to emphasize their dependence on the corresponding datasets $\mathcal{D}_{n}^{(\cdot)}$, such that
\begin{align}
    \widehat{\bm{\rho}}_{\vect{\omega},n} &= 
    \widehat{\y}_{(\vect{\omega})}\,\big|\, 
    \overline{\delta \q},\,\mathcal{D}_{n}^{(\vect{\omega})}, 
    \label{eq:gp_rho_w}\\
    \widehat{\bm{\rho}}_{\vect{v},n} &=
    \widehat{\y}_{(\vect{v})}\,\big|\, 
    \delta \Q,\,\mathcal{D}_{n}^{(\vect{v})}, 
    \label{eq:gp_rho_v}
\end{align}
where $\widehat{\y}_{(\vect{\omega})}\in \mathbb{H}$ and 
$\widehat{\y}_{(\vect{v})}\in \mathcal{H}$ denote the GP estimates of the target variables 
$\widetilde{\y}_{(\vect{\omega})}$ and $\widetilde{\y}_{(\vect{v})}$, respectively.

This framework naturally supports both \textit{offline learning} (using previously collected data only) 
and hybrid \textit{online/offline} schemes, where new samples can be continuously incorporated or 
selectively forgotten depending on the learning strategy.

\begin{assumption}\label{assumption:1}
    The number of datasets $\mathcal{D}_{n}^{(\vect{\omega})}, \mathcal{D}_{n}^{(\vect{v})}$ is finite and there are only finite many switches of $n(t)$ over time, such that there exists a time $T_{end}\in \mathbb{R}_{\geq 0}$ where $n(t)={n_{end}}$, $\forall t\geq T_{end}$. 
\end{assumption}

Note that Assumption \ref{assumption:1} is little restrictive since the number of sets is often naturally bounded due to finite computational power or memory limitations and, since the unknown functions $\bm{\rho}_{\vect{\omega}}$, $\bm{\rho}_{\vect{v}}$ in \eqref{omegabarerror} and \eqref{verror} are not explicitly time-dependent, long-life learning is typically not required. Therefore, there exist constant datasets $\mathcal{D}_{n_{end}}^{(\vect{\omega})}, \mathcal{D}_{n_{end}}^{(\vect{v})}$ for all $t>T_{end}$. Furthermore, Assumption \ref{assumption:1} ensures that the switching between the datasets is not infinitely fast, which is natural in real-world applications. Note also that, in practice, since missions are finite in time, we rely on the assumption that the time horizon is sufficiently long for $\D_{n_{end}}^{(\vect{\omega})}$ and $\D_{n_{end}}^{(\vect{v})}$ to be meaningful.


Gaussian Process (GP) models are powerful oracles for nonlinear function regression. 
For the prediction step, we concatenate the $N(n)$ training points of each dataset 
$\mathcal{D}_{n}^{(\cdot)}$ into input matrices and corresponding output matrices as
\begin{align}
    X^{(\vect{\omega})} &= 
    \big[\overline{\delta \q}^{\{1\}},\overline{\delta \q}^{\{2\}},\dots, 
    \overline{\delta \q}^{\{N(n)\}}\big]^{T} \in \mathbb{H}^{N(n)}, \\
    X^{(\vect{v})} &= 
    \big[\delta \Q^{\{1\}},\delta \Q^{\{2\}},\dots,
    \delta \Q^{\{N(n)\}}\big]^{T} \in \mathcal{H}^{N(n)}, \\
    Y^{(\vect{\omega})} &= 
    \big[\widehat{\y}_{(\vect{\omega})}^{1},\widehat{\y}_{(\vect{\omega})}^{2},\dots,
    \widehat{\y}_{(\vect{\omega})}^{N(n)}\big], \\
    Y^{(\vect{v})} &= 
    \big[\widehat{\y}_{(\vect{v})}^{1},\widehat{\y}_{(\vect{v})}^{2},\dots,
    \widehat{\y}_{(\vect{v})}^{N(n)}\big],
\end{align}
where the measurements 
$\widehat{\y}_{(\vect{\omega})}=\widetilde{\y}_{(\vect{\omega})}+\eta$ and 
$\widehat{\y}_{(\vect{v})}=\widetilde{\y}_{(\vect{v})}+\eta$ 
are corrupted by additive Gaussian noise 
$\eta\sim\mathcal{N}(0,\sigma^{2}_{\nu})$.

\paragraph*{GP prediction.}
Let $\overline{\delta \q}^{\dagger}\in\mathbb{H}$ and 
$\delta\Q^{\dagger}\in\mathcal{H}$ be new test points. 
The corresponding GP predictions 
$\y^{\dagger}_{(\vect{\omega})}\in\mathbb{H}$ and 
$\y^{\dagger}_{(\vect{v})}\in\mathcal{H}$ are given by
\begin{align}
    \mu(\y^{\dagger}_{(\cdot)} \vert x^{\dagger}, \mathcal{D}^{(\cdot)}) 
    &= m^{(\cdot)}(x^{\dagger})
    + \bm{k}^{(\cdot)}(x^{\dagger}, X^{(\cdot)})^{T} 
    K_{(\cdot)}^{-1} Y_{:,i}^{(\cdot)}, \label{eq:gp_mean_pred}\\
    \text{var}(\y^{\dagger}_{(\cdot)} \vert x^{\dagger}, \mathcal{D}^{(\cdot)}) 
    &= k^{(\cdot)}(x^{\dagger},x^{\dagger})
    \nonumber\\&- \bm{k}^{(\cdot)}(x^{\dagger}, X^{(\cdot)})^{T}
    K_{(\cdot)}^{-1}\bm{k}^{(\cdot)}(x^{\dagger}, X^{(\cdot)}), 
    \label{eq:gp_var_pred}
\end{align}
for all $i\in\{1,2,3\}$, where $Y_{:,i}^{(\cdot)}$ denotes the $i$-th column of the 
output matrix $Y^{(\cdot)}$. 
The mean function $m^{(\cdot)}(\cdot)$ encodes prior knowledge of the system, 
while the kernel $k^{(\cdot)}(\cdot,\cdot)$ defines input correlations in either 
$\mathbb{H}$ or $\mathcal{H}$, depending on the modeled variable.

\paragraph*{Kernel and covariance definitions}
The Gram matrix $K_{(\cdot)} \in \mathbb{R}^{N(n)\times N(n)}$ is defined elementwise as
\begin{equation}
    K_{j',j}^{(\cdot)} = 
    k^{(\cdot)}(X^{(\cdot)}_{:,j'},X^{(\cdot)}_{:,j})
    + \delta(j,j')\,\sigma^{2}, 
    \, j,j'\in\{1,\dots,N(n)\},
\end{equation}
where $\delta(j,j')=1$ if $j=j'$ and $0$ otherwise. 
The vector-valued covariance function 
$\bm{k}^{(\cdot)}:\mathcal{X}\times\mathcal{X}\to\mathbb{R}^{N(n)}$ 
(with $\mathcal{X}\in\{\mathbb{H},\mathcal{H}\}$) 
has elements 
$k^{(\cdot)}_{j}=k^{(\cdot)}(x^{\dagger},X^{(\cdot)}_{:,j})$ 
for $j=1,\dots,N(n)$, expressing the covariance between the test input $x^{\dagger}$ 
and the training data $X^{(\cdot)}$.

\begin{remark}\label{remark2}
Unit quaternions $\bar{\boldsymbol q},\bar{\boldsymbol q}' \in \mathbb S^3 \subset \mathbb R^4$
double-cover $\mathrm{SO}(3)$, hence $\bar{\boldsymbol q}$ and $-\bar{\boldsymbol q}$ represent the same rotation.
To avoid learning duplicated representations, kernels on quaternion inputs should be invariant under the antipodal map
$\bar{\boldsymbol q}\mapsto -\bar{\boldsymbol q}$.
A common geometry-consistent choice is the antipodally invariant chordal distance
\[
d_{\pm}(\bar{\boldsymbol q},\bar{\boldsymbol q}') := \min\{\|\bar{\boldsymbol q}-\bar{\boldsymbol q}'\|_2,\ \|\bar{\boldsymbol q}+\bar{\boldsymbol q}'\|_2\}
= \sqrt{2-2|\langle \bar{\boldsymbol q},\bar{\boldsymbol q}'\rangle|},
\]
where $\|\cdot\|_2$ and $\langle\cdot,\cdot\rangle$ denote the Euclidean norm and inner product in $\mathbb R^4$.
Composing $d_{\pm}$ with a standard positive-definite kernel (e.g., squared exponential) yields an antipodally invariant kernel,
e.g.,
\[
k^{(\boldsymbol\omega)}(\bar{\boldsymbol q},\bar{\boldsymbol q}') :=
\sigma_f^2\exp\!\Big(-\tfrac{d_{\pm}(\bar{\boldsymbol q},\bar{\boldsymbol q}')^2}{2\ell^2}\Big),
\] where \(\sigma_f^2\) denotes the signal variance and \(\ell_d\) are tje dimension-wise lenghscales, which has been used for GP learning of rotations; see \cite{lang2015gaussian}.

This construction extends naturally to rigid-body motions represented by unit dual quaternions
$\boldsymbol Q, \boldsymbol Q'\in \mathcal{H}_u \subset \mathbb R^8$,
which double-cover $\mathrm{SE}(3)$ (i.e., $\boldsymbol Q$ and $-\boldsymbol Q$ represent the same pose).
Let $\boldsymbol Q\leftrightarrow(\bar{\boldsymbol q},\boldsymbol p)$ and
$\boldsymbol Q'\leftrightarrow(\bar{\boldsymbol q}',\boldsymbol p')$, with $\bar{\boldsymbol q},\bar{\boldsymbol q}'\in\mathbb S^3$ and
$\boldsymbol p,\boldsymbol p'\in\mathbb R^3$.
A geometry-aware distance on $\mathrm{SE}(3)$ can be defined, for instance, by
\[
d_{\mathrm{SE}(3)}(\boldsymbol Q, \boldsymbol Q') :=
\sqrt{\,d_{\pm}(\bar{\boldsymbol q},\bar{\boldsymbol q}')^2 + \tfrac{1}{\lambda^2}\|\boldsymbol p-\boldsymbol p'\|_2^2\,},
\]
where $\lambda>0$ balances translational vs rotational scales.
Then
\[
k^{(\boldsymbol v)}(\boldsymbol Q,\boldsymbol Q') :=
\sigma_f^2\exp\!\Big(-\tfrac{d_{\mathrm{SE}(3)}(\boldsymbol Q,\boldsymbol Q')^2}{2\ell^2}\Big)
\]
is invariant under $\boldsymbol Q\mapsto -\boldsymbol Q$ and respects the group geometry.
Related geometry-aware GP constructions on Lie groups and their use in learning-based control are discussed in
\cite{lang2015gaussian,beckers2021online}.
\end{remark}

\medskip

\paragraph*{Model selection and hyperparameters}
The choice of kernel and the determination of its hyperparameters 
constitute the degrees of freedom of the regression. 
The hyperparameters, as well as the Gaussian noise variance $\sigma^{2}_{\nu}$, 
are estimated by maximizing the marginal log-likelihood 
(see~\cite{rasmussen2006gaussian}). 
A widely used kernel for GP models of physical systems—and the one adopted in our experiments—
is the \textit{squared exponential kernel} with isotropic distance measure. 
An overview of alternative kernel properties can be found in~\cite{rasmussen2006gaussian}.

The mean functions $m^{(\vect{\omega})}$ and $m^{(\vect{v})}$ 
can be obtained from standard system identification techniques 
for the unknown dynamics $\vect{\rho_\omega}$ and $\vect{\rho_v}$ 
(see~\cite{aastrom1971system}). 
However, in the absence of prior knowledge, they are set to zero, i.e.,
\[
m^{(\vect{\omega})}(\overline{\delta \q}^{\dagger})=0, 
\qquad
m^{(\vect{v})}(\delta\Q^{\dagger})=0.
\]

\paragraph*{Multivariate formulation}
Based on \eqref{eq:gp_mean_pred}--\eqref{eq:gp_var_pred}, the three output components define a multivariate Gaussian
distribution with mean vector and covariance matrix given by the stacked posterior means and variances across the
three scalar GPs. Likewise, the corresponding distribution for 
$\y^{\dagger}_{(\vect{v})}\vert x^{\dagger}$ and $\mathcal{D}_{n}^{(\vect{v})}$ 
is denoted by 
$\mu(\y^{\dagger}_{(\vect{v})}\vert x^{\dagger},\mathcal{D}_{n}^{(\vect{v})})$ and 
$\Sigma(\y^{\dagger}_{(\vect{v})}\vert x^{\dagger},\mathcal{D}_{n}^{(\vect{v})})$.


\begin{remark}
For simplicity, identical kernels are considered for all output dimensions. 
Nevertheless, the GP model can be easily extended to use distinct kernels for each output dimension. 
Since the GP is applied in an online setting where new data are continuously collected, 
the datasets $\mathcal{D}^{(\vect{\omega})}_n$ and $\mathcal{D}^{(\vect{v})}_n$ used for prediction 
(see~\eqref{eq:gp_mean_pred}–\eqref{eq:gp_var_pred}) evolve over time. 
The GP framework naturally accommodates new training samples, 
as any subset of data follows a multivariate Gaussian distribution. 
For details on online learning performance, see~\cite{dai2024cooperative} and~\cite{dai2024decentralized}.
\end{remark}

Assume the measurement noises in \eqref{eq:gp_dataset} are i.i.d.\ and independent of the inputs,
and training samples are collected such that standard GP concentration results apply
(see \cite{srinivas2012information}).

\begin{assumption}\label{assumption:2}
Consider the Gaussian Process predictions 
$\widehat{\bm{\rho}}_{\vect{\omega},n}\in\mathcal{C}^{0}$ and 
$\widehat{\bm{\rho}}_{\vect{v},n}\in\mathcal{C}^{0}$ based on the datasets 
$\mathcal{D}_{n}^{(\vect{\omega})}$ and $\mathcal{D}_{n}^{(\vect{v})}$.
Let $Q^{(\vect{\omega})}_{\mathcal{X}}\subset\mathbb{H}$ 
and $Q^{(\vect{v})}_{\mathcal{X}}\subset\mathcal{H}$ 
be compact subsets where $\widehat{\bm{\rho}}_{\vect{\omega},n}$ 
and $\widehat{\bm{\rho}}_{\vect{v},n}$ are bounded, respectively.  
Then, there exist finite bounding functions
\[
\bm{\rho}^{\vect{\omega},\ddagger}_{n}:Q^{(\vect{\omega})}_{\mathcal{X}}\to\mathbb{R}_{\geq 0}, 
\qquad 
\bm{\rho}^{\vect{v},\ddagger}_{n}:Q^{(\vect{v})}_{\mathcal{X}}\to\mathbb{R}_{\geq 0},
\]
such that the prediction errors satisfy, for all 
$\overline{\delta\q}\in Q^{(\vect{\omega})}_{\mathcal{X}}$, 
$\delta\Q\in Q^{(\vect{v})}_{\mathcal{X}}$, and 
$n\in\{1,\dots,n_{\text{end}}\}$,
\begin{align}
    P\!\left(
    \big\|\bm{\rho}_{\vect{\omega}} - 
    \widehat{\bm{\rho}}_{\vect{\omega},n}\big\| 
    \leq 
    \bm{\rho}^{\vect{\omega},\ddagger}_{n}(\overline{\delta\q},\gamma)
    \right)
    &\geq \gamma_{\vect{\omega}}, 
    \label{eq:probab_w}\\[4pt]
    P\!\left(
    \big\|\bm{\rho}_{\vect{v}} - 
    \widehat{\bm{\rho}}_{\vect{v},n}\big\| 
    \leq 
    \bm{\rho}^{\vect{v},\ddagger}_{n}(\delta\Q,\gamma)
    \right)
    &\geq \gamma_{\vect{v}},
    \label{eq:probab_v}
\end{align}
where $\gamma_{\vect{\omega}},\gamma_{\vect{v}}\in(0,1]$ denote the corresponding confidence probabilities.
\end{assumption}

\begin{remark}
Assumption~\ref{assumption:2} ensures that for each dataset $\mathcal{D}_{n}$ 
there exists a probabilistic upper bound on the GP prediction error. 
Specifically, it bounds the difference between the actual dynamics 
$\bm{\rho}_{\vect{\omega}}(\overline{\delta\q})$ and its estimate 
$\widehat{\bm{\rho}}_{\vect{\omega},n}(\overline{\delta\q})$ 
over $Q^{(\vect{\omega})}_{\mathcal{X}}$, 
and analogously for 
$\bm{\rho}_{\vect{v}}(\delta\Q)$ and 
$\widehat{\bm{\rho}}_{\vect{v},n}(\delta\Q)$ 
over $Q^{(\vect{v})}_{\mathcal{X}}$.
\end{remark}

To provide model error bounds, 
additional assumptions on the unknown components of 
$\mathcal{D}_{n}^{(\vect{\omega})}$ and 
$\mathcal{D}_{n}^{(\vect{v})}$ 
must be introduced~\cite{wolpert1996lack}.

\begin{assumption}\label{assumption:3}
The kernel $k$ is selected such that 
$\bm{\rho}_{\vect{\omega}}$ and $\bm{\rho}_{\vect{v}}$ 
have finite reproducing kernel Hilbert space (RKHS) norms 
on $Q^{(\vect{\omega})}_{\mathcal{X}}$ and 
$Q^{(\vect{v})}_{\mathcal{X}}$, respectively, i.e.,
\[
\|\bm{\rho}_{\vect{\omega}}(\overline{\delta\q})\|_{k}=\xi_1<\infty,
\qquad
\|\bm{\rho}_{\vect{v}}(\delta\Q)\|_{k}=\xi_2<\infty.
\]
\end{assumption}

\begin{remark}
Assumption~\ref{assumption:3} is satisfied, for instance, by universal kernels constructed
from geometry-aware and antipodally invariant distance functions on
$\mathrm{SO}(3)$ and $\mathrm{SE}(3)$.
Such kernels are obtained by composing standard positive-definite kernels
(e.g., squared exponential kernels) with smooth, geometry-consistent distances
that respect the double-cover structure of unit quaternions and unit dual quaternions.

Since $\mathrm{SO}(3)$ and $\mathrm{SE}(3)$ are compact when restricted to the sets
$Q^{(\vect{\omega})}_{\mathcal{X}}$ and $Q^{(\vect{v})}_{\mathcal{X}}$,
the resulting kernels remain universal and induce reproducing kernel Hilbert spaces
that are dense in the space of continuous functions on these sets.
Consequently, Assumption~\ref{assumption:3} is not restrictive in practice,
as any continuous unknown dynamics
$\bm{\rho}_{\vect{\omega}}$ and $\bm{\rho}_{\vect{v}}$
can be approximated arbitrarily well while preserving the underlying group symmetries.
\end{remark}

\smallskip

The boundedness of the RKHS norm is related to the properties of the unknown functions, 
rather than the kernel itself, although the kernel choice influences the numerical value of the bound. 
Assumption~\ref{assumption:3} requires that the selected kernel ensures 
$\bm{\rho}_{\vect{\omega}}$ and $\bm{\rho}_{\vect{v}}$ belong to the associated RKHS. 
This assumption may appear restrictive at first sight, since the unknown functions are not known explicitly. However, there exist universal kernels that can approximate any continuous function arbitrarily well on compact sets~\cite[Lemma~4.55]{steinwart2008support}.
When such kernels are constructed using geometry-consistent and antipodally invariant
distances on $\mathrm{SO}(3)$ and $\mathrm{SE}(3)$, as discussed above,
the resulting RKHS naturally respects the group structure while retaining
the universal approximation property. Using such kernels, the bounded RKHS norm error can be constrained as stated in the following lemma.

\begin{lemma}[adapted from~\cite{srinivas2012information}]\label{lemma:3}
Consider the unknown functions 
$\bm{\rho}_{(\cdot)}\in\{\bm{\rho}_{\vect{\omega}},\bm{\rho}_{\vect{v}}\}$ 
and the GP models satisfying Assumption~\ref{assumption:3}. 
Then, for all 
$x\in Q^{(\cdot)}_{\mathcal{X}}$, 
$\gamma_{(\cdot)}\in(0,1)$, 
and $n\in\{1,\dots,n_{\text{end}}\}$, 
the model error satisfies
\begin{scriptsize}\begin{equation}
    P\!\left(
    \big\|
    \mu(\widehat{\bm{\rho}}_{(\cdot),n}\,|\,x,\mathcal{D}_{n}^{(\cdot)})
    - \bm{\rho}_{(\cdot)}
    \big\|
    \leq 
    \big\|
    \bm{\beta}_{(\cdot),n}^{T}
    \Sigma^{1/2}\!
    \left(
    \widehat{\bm{\rho}}_{(\cdot),n}\,|\,x,\mathcal{D}_{n}^{(\cdot)}
    \right)
    \big\|
    \right)
    \geq \gamma_{(\cdot)}.
    \label{eq:lemma_compact}
\end{equation}\end{scriptsize}
Here $\bm{\beta}_{(\cdot),n}\in\mathbb{H}$ is defined elementwise as
\begin{equation}
    (\bm{\beta}_{(\cdot),n})_{j} =
    \sqrt{
    2\|\bm{\rho}_{(\cdot),n}\|_{k}^{2}
    + 300\,\Gamma_{j}^{(\cdot)}
    \ln^{3}\!\left(
    \frac{N(n)+1}{1-\gamma_{(\cdot)}^{1/3}}
    \right)},
    \label{eq:beta_compact}
\end{equation}
where for $x,x'\in\{x^{\{1\}},\dots,x^{\{N(n)+1\}}\}$ the \textit{maximum information gain} term $\Gamma_{j}^{(\cdot)}\in\R$ is given by
\begin{equation}
    \Gamma^{(\cdot)}_{j} =
    \max_{x^{\{1\}},\dots,x^{\{N(n)+1\}}\in Q^{(\cdot)}_\mathcal{X}}
    \frac{1}{2}
    \log
    \!\left|
    I+\sigma_{j}^{-2}
    K_{(\cdot)}(x,x')
    \right|.
    \label{eq:gamma_compact}
\end{equation} 
\end{lemma}

\textit{Proof:} 
It follows directly from~\cite[Theorem~6]{srinivas2012information}. 
\hfill$\square$

Consequently, the probabilistic error bounds from Assumption~\ref{assumption:2} can be written compactly as
\begin{equation}
    \bm{\rho}^{(\cdot),\ddagger}_{n}(x,\gamma_{(\cdot)}) :=
    \big\|
    \bm{\beta}_{(\cdot),n}^{T}
    \Sigma^{1/2}\!
    \left(
    \widehat{\bm{\rho}}_{(\cdot),n}\,|\,x,\mathcal{D}_{n}^{(\cdot)}
    \right)
    \big\|.
    \label{eq:rho_ddagger_compact}
\end{equation}

\begin{remark}
An efficient algorithm can compute $\bm{\beta}_{n}$ based on the maximum information gain. 
Although the entries of $\bm{\beta}_{n}$ typically increase with the amount of training data, 
the true functions $\bm{\rho}_{\vect{\omega}}$ and $\bm{\rho}_{\vect{v}}$ 
can be learned with arbitrarily small error due to the decreasing predictive variance $\Sigma$ 
(see~\cite{Berkenkamp2016ROA}). 
In general, the prediction error bounds 
$\bm{\rho}^{\vect{\omega},\ddagger}_{n}(\overline{\delta\q},\gamma_{\vect{\omega}})$ and 
$\bm{\rho}^{\vect{v},\ddagger}_{n}(\delta\Q,\gamma_{\vect{v}})$ 
are larger when GP prediction uncertainty is high, and smaller otherwise. 
The bounds also tend to increase as the compact sets 
$Q^{(\vect{\omega})}_\mathcal{X}$ and $Q^{(\vect{v})}_\mathcal{X}$ expand. 
The stochastic nature of the bound arises from the finite number of noisy training samples.
\end{remark}

\begin{remark}
    The probabilistic bound given in Lemma \ref{lemma:3} can be compared with others in the recent literature in terms of computational complexity, since the training becomes sometimes intractable as data grows without a bound. In particular, these scalability challenges to adaptively process streaming data in real time have been recently studied in \cite{lederer2019uniform}, \cite{lederer2021gaussian}, \cite{lederer2024safe}, \cite{yang2025streaming}.

In this work, we use an adaptation of the bound given in \cite{srinivas2012information}, which is currently the most used in the literature for trajectory tracking of mechanical systems (see, for instance \cite{beckers2019stable}, \cite{umlauft2019feedback}, \cite{capone2019backstepping},  \cite{beckers2021online}, \cite{beckers2023data}, \cite{beckers2025physics}). Despite a comparative analysis with different performance methods, it is not a topic to study in this paper; one could compare the efficiency of such bounds in the class of systems we study in this work in a further paper. 
\end{remark}

\subsection{Control design}

We now design a learning-based pose controller for the disturbed system
\eqref{omegabarerror}–\eqref{verror}. The idea is to start from the nominal
controller \eqref{omegabar}–\eqref{vbar} and compensate the unknown terms
$\vect{\rho}_{\vect{\omega}}$ and $\vect{\rho}_{\vect{v}}$ using the GP
predictions. The stability analysis is carried out with a family of Lyapunov
functions $\{V_n\}$, where the $n$-th function is active when the GP uses
the corresponding datasets $\mathcal{D}^{(\vect{\omega})}_n$ and
$\mathcal{D}^{(\vect{v})}_n$ for prediction. By Assumption~\ref{assumption:1},
the number of switches is finite, so switching between stable subsystems cannot
generate unbounded trajectories~\cite{liberzon1999basic}.

\begin{theorem}\label{thm:lbc}
Consider the dual-quaternion kinematics \eqref{quat-dual-dynamic} and let
$\widetilde{\vect{\omega}}_d = (\vect{\omega}_d,0)$ and
$\widetilde{\vect{v}}_d = (\vect{v}_d,0)\in\mathbb{H}$, and a desired dual
quaternion $\Q_d$ satisfying \eqref{vehic-dyn}.  

Assume that the actual inputs are perturbed as in \eqref{omegabarerror}–\eqref{verror},
with unknown functions $\vect{\rho}_{\vect{\omega}}$ and $\vect{\rho}_{\vect{v}}$
modeled by Gaussian Process (GP) priors satisfying
Assumptions~\ref{assumption:1}–\ref{assumption:3}
and the probabilistic error bounds of Lemma~\ref{lemma:3}, with confidence
levels $\gamma_{\vect{\omega}},\gamma_{\vect{v}}\in(0,1]$. Let the gain matrices
$\K_{\vect{\omega},n},\K_{\vect{v},n}:\mathbb{R}\to\mathbb{R}^{3\times 3}$
be uniformly bounded, symmetric and positive definite.  
That is, there exist constants $\alpha_n^{(\vect{\omega})}>0$, and  $\alpha_n^{(\vect{v})}>0$ such that, for all arguments $\beta$, $\K_{\vect{\omega},n}(\beta)\succeq\alpha_n^{(\vect{\omega})} I$, $\K_{\vect{v},n}(\beta)\succeq\alpha_n^{(\vect{v})} I$. For compactness, define $\alpha_n := \min\{\alpha_n^{(\vect{\omega})},\,\alpha_n^{(\vect{v})}\}$.

Assume that $\Q$ evolves according to \eqref{vehic-dyn} with the nominal
control laws \eqref{omegabar}–\eqref{vbar}, and define the learned controllers
\begin{align}
\widetilde{\vect{\omega}}_{c}
&= \widetilde{\vect{\omega}}
   - \mu\!\left(
        \widehat{\bm{\rho}}_{\vect{\omega},n}
        \,\big|\,
        \overline{\delta \q},\mathcal{D}_{n}^{(\vect{\omega})}
     \right),
\label{learnedcontrol}\\[1mm]
\widetilde{\vect{v}}_{c}
&= \widetilde{\vect{v}}
   - \mu\!\left(
        \widehat{\bm{\rho}}_{\vect{v},n}
        \,\big|\,
        \delta \Q,\mathcal{D}_{n}^{(\vect{v})}
     \right),
\label{learnedcontrolv}
\end{align} where 
$\overline{\delta \q}=\PrincipalPart(\delta \Q)$, 
$\delta \Q = \Q_d^*\circ \Q$, 
and $\mu(\cdot|\cdot,\mathcal{D}_{n}^{(\cdot)})$ denotes the GP posterior mean
given the dataset $\mathcal{D}_{n}^{(\cdot)}$. With \eqref{omegabarerror}--\eqref{verror}, the closed-loop inputs become
$\widetilde{\vect{\omega}}_\rho=\widetilde{\vect{\omega}}_{c}+\bm{\rho}_{\vect{\omega}}(\overline{\delta\q})+\nu_\omega$
and
$\widetilde{\vect{v}}_\rho=\widetilde{\vect{v}}_{c}+\bm{\rho}_{\vect{v}}(\delta\Q)+\nu_v$,
so that the residual perturbations are exactly the GP model errors.

For each $n$, define
\begin{align*}
c^{(\vect{\omega})}_n
&:= \max_{\overline{\delta \q}\in Q^{(\vect{\omega})}_{\mathcal{X}}}
        \bm{\rho}^{(\vect{\omega}),\ddagger}_{n}(\overline{\delta \q},\gamma_{\vect{\omega}}),\\[1mm]
c^{(\vect{v})}_n
&:= \frac{1}{2\alpha_n^{(\vect{v})}}
        \max_{\delta \Q\in Q^{(\vect{v})}_{\mathcal{X}}}
        \big(
            \bm{\rho}^{(\vect{v}),\ddagger}_{n}(\delta \Q,\gamma_{\vect{v}})
        \big)^{2},
\end{align*}
and set
\[
\varepsilon_{0,n} := \frac{c^{(\vect{\omega})}_n + c^{(\vect{v})}_n}{\alpha_n}.
\]
Finally define $K_{\varepsilon_{0,n}}
:= \left\{
      \delta \Q \;:\;
      |\delta q_0|\,\|\delta \q\|^{2}
      + \|\delta \p\|^{2}
      \le \varepsilon_{0,n}
   \right\}$, and $M_n := \max_{\delta \Q \in K_{\varepsilon_{0,n}}} V_n(\delta \Q)$.

Then, there exist a time $T\ge 0$ and an index $n_{\text{end}}$
(as in Assumption~\ref{assumption:1}) such that, with $\gamma := \min\{\gamma_{\vect{\omega}},\,\gamma_{\vect{v}}\}$, the dual-quaternion error
$\delta \Q$ is uniformly ultimately bounded in probability:
\begin{equation}
    P\left\{
      \|\delta \q(t)\|^{2}
      + \frac{1}{2}\|\delta \p(t)\|^{2}
      \le M_{n_{\text{end}}},
      \ \forall t\ge T
    \right\}
    \ge \gamma.
    \label{modelerror}
\end{equation}
\end{theorem}

\textit{Proof:}
Consider, for each $n$, the Lyapunov candidate
\begin{equation}
    V_n(\delta \Q)
    = \|\delta \q\|^{2}
      + \frac{1}{2}\|\delta \p\|^{2},
\end{equation}
which is positive definite in $\delta \Q$ and satisfies
$V_n(\delta \Q)=0$ if and only if
$\delta \q = \vect{0}$, $|\delta q_0|=1$ and
$\DualPart(\delta\Q)=(\vect{0},0)$.

Taking the time derivative of $V_n$ along the solutions of the
error dynamics $\dot{\delta \Q}$ (see \eqref{quat-dual-dynamic}) we obtain
\begin{equation}\label{eq:Vdot_general}
    \dot{V}_n
    = \delta \q^{T} \big(\mathbf{S} (\delta \q)+I \delta q_{0}\big)\,\mathcal{P} (\delta \boldsymbol{\Omega})
      + \delta \p^{T} \dot{\delta \p},
\end{equation}
where $\mathcal{P}(\delta \bf{\Omega})$ denotes the primary (non-dual) part of the
dual quaternion $\delta\bf{\Omega}$.

For the nominal controller \eqref{omegabar}–\eqref{vbar} (without
unknown perturbations and without GP compensation), it is standard
(see Theorem~1) that the error dynamics yield
\begin{equation}\label{eq:Vdot_nominal}
    \dot{V}_n
    = - |\delta q_0| \,\delta \q^T \K_{\vect{\omega},n}(\beta)\,\delta \q
      - \delta \p^T \K_{\vect{v},n}(\beta)\,\delta \p.
\end{equation}
Using the learned controllers \eqref{learnedcontrol}–\eqref{learnedcontrolv},
we can write the closed-loop perturbation of \eqref{eq:Vdot_nominal} as
additional terms depending on the GP model errors.

Introduce the GP model error functions
\begin{align}
    \e_{\vect{\omega},n}(\overline{\delta \q})
    &:= \vect{\rho}_{\vect{\omega},n}(\overline{\delta \q})
        - \mu\!\left(
              \widehat{\vect{\rho}}_{\vect{\omega},n}
              \,\big|\,
              \overline{\delta \q},\mathcal{D}_{n}^{(\vect{\omega})}
           \right),\\[1mm]
    \e_{\vect{v},n}(\delta \Q)
    &:= \vect{\rho}_{\vect{v},n}(\delta \Q)
        - \mu\!\left(
              \widehat{\vect{\rho}}_{\vect{v},n}
              \,\big|\,
              \delta \Q,\mathcal{D}_{n}^{(\vect{v})}
           \right).
\end{align}
Then the derivative \eqref{eq:Vdot_general}, with the learned controllers
\eqref{learnedcontrol}–\eqref{learnedcontrolv}, can be expressed as
\begin{align}
    \dot{V}_n
    &= - |\delta q_0| \,\delta \q^T \K_{\vect{\omega},n}(\beta)\,\delta \q
       - \delta \p^T \K_{\vect{v},n}(\beta)\,\delta \p \nonumber\\
    &\quad
       - \delta \q^{T}\big[\mathbf{S}(\delta \q)+ I\delta q_{0}\big]\,
         \e_{\vect{\omega},n}(\overline{\delta \q})
       - \delta \p^{T}\e_{\vect{v},n}(\delta \Q).
       \label{eq:Vdot_full}
\end{align}

Using that $\mathbf{S}(\delta \q)$ is skew-symmetric, we have
$\delta \q^T\mathbf{S}(\delta \q)=0$, and therefore
$
\delta \q^{T}\big[\mathbf{S}(\delta \q)+ I\delta q_{0}\big]
= \delta q_0\,\delta \q^{T}$.
Hence \eqref{eq:Vdot_full} simplifies to
\begin{align}
    \dot{V}_n
    &= - |\delta q_0| \,\delta \q^T \K_{\vect{\omega},n}(\beta)\,\delta \q
       - \delta \p^T \K_{\vect{v},n}(\beta)\,\delta \p \nonumber\\
    &\quad
       - \delta q_0\,\delta \q^{T}\e_{\vect{\omega},n}(\overline{\delta \q})
       - \delta \p^{T}\e_{\vect{v},n}(\delta \Q).
       \label{eq:Vdot_full_simple}
\end{align}

By the gain conditions, $\K_{\vect{\omega},n}(\beta)\succeq\alpha_n^{(\vect{\omega})} I$ and $\K_{\vect{v},n}(\beta)\succeq\alpha_n^{(\vect{v})} I$, so
\begin{equation}\label{eq:Vdot_gain_bound}
    - |\delta q_0| \,\delta \q^T \K_{\vect{\omega},n}(\beta)\,\delta \q
    \le - \alpha_n^{(\vect{\omega})} |\delta q_0| \|\delta \q\|^{2},
\end{equation}
and
\begin{equation}\label{eq:Vdot_gain_bound_v}
    - \delta \p^T \K_{\vect{v},n}(\beta)\,\delta \p
    \le - \alpha_n^{(\vect{v})} \|\delta \p\|^{2}.
\end{equation}

Next we bound the two GP error terms. Since $\overline{\delta \q}$ is a unit
quaternion, we have $\|\delta \q\|\le 1$ and $|\delta q_0|\le 1$, and thus
\begin{align}
    \big|
    \delta q_0\,\delta \q^{T}\e_{\vect{\omega},n}(\overline{\delta \q})
    \big|
    &\le \|\delta \q\|\,
         |\delta q_0|\,
         \|\e_{\vect{\omega},n}(\overline{\delta \q})\|
     \le \|\e_{\vect{\omega},n}(\overline{\delta \q})\|.
     \label{eq:bound_omega_term}
\end{align}
For the translational part we apply Cauchy–Schwarz and the inequality
$2ab\le a^{2}+b^{2}$:
\begin{align}
    \big|
    \delta \p^{T}\e_{\vect{v},n}(\delta \Q)
    \big|
    &\le \|\delta \p\|\,
         \|\e_{\vect{v},n}(\delta \Q)\| \nonumber\\
    &\le \tfrac{1}{2}\alpha_n^{(\vect{v})}\|\delta \p\|^{2}
         + \frac{1}{2\alpha_n^{(\vect{v})}}
           \|\e_{\vect{v},n}(\delta \Q)\|^{2}.
    \label{eq:bound_v_term}
\end{align}

Substituting \eqref{eq:Vdot_gain_bound}–\eqref{eq:bound_v_term}
into \eqref{eq:Vdot_full_simple} yields
\begin{align}
    \dot{V}_n
    &\le - \alpha_n^{(\vect{\omega})} |\delta q_0| \|\delta \q\|^{2}
         - \alpha_n^{(\vect{v})} \|\delta \p\|^{2} \nonumber\\
    &\quad
         + \|\e_{\vect{\omega},n}(\overline{\delta \q})\|
         + \tfrac{1}{2}\alpha_n^{(\vect{v})}\|\delta \p\|^{2}
         + \frac{1}{2\alpha_n^{(\vect{v})}}
           \|\e_{\vect{v},n}(\delta \Q)\|^{2} \nonumber\\[1mm]
    &= - \alpha_n^{(\vect{\omega})} |\delta q_0| \|\delta \q\|^{2}
       - \tfrac{1}{2}\alpha_n^{(\vect{v})} \|\delta \p\|^{2} \nonumber\\
    &\quad
       + \|\e_{\vect{\omega},n}(\overline{\delta \q})\|
       + \frac{1}{2\alpha_n^{(\vect{v})}}
         \|\e_{\vect{v},n}(\delta \Q)\|^{2}.
       \label{eq:Vdot_pre_prob}
\end{align}

By Assumption~\ref{assumption:2} and Lemma~\ref{lemma:3},
for each $n\in\{1,\dots,n_{\mathrm{end}}\}$ and for all
$\overline{\delta \q}\in Q^{(\vect{\omega})}_{\mathcal{X}}$,
$\delta \Q\in Q^{(\vect{v})}_{\mathcal{X}}$, the GP prediction errors satisfy
\begin{align}
    P\!\left(
      \|\e_{\vect{\omega},n}(\overline{\delta \q})\|
      \le
      \bm{\rho}^{(\vect{\omega}),\ddagger}_{n}(\overline{\delta \q},\gamma_{\vect{\omega}})
    \right)
    &\ge \gamma_{\vect{\omega}}, \label{eq:prob_bound_omega}\\[1mm]
    P\!\left(
      \|\e_{\vect{v},n}(\delta \Q)\|
      \le
      \bm{\rho}^{(\vect{v}),\ddagger}_{n}(\delta \Q,\gamma_{\vect{v}})
    \right)
    &\ge \gamma_{\vect{v}}. \label{eq:prob_bound_v}
\end{align}
Let $\gamma := \min\{\gamma_{\vect{\omega}},\gamma_{\vect{v}}\}\in(0,1]$, and consider the event where both bounds
\eqref{eq:prob_bound_omega}–\eqref{eq:prob_bound_v} hold.
On this event (which has probability at least $\gamma$) we have
\begin{align}
    \|\e_{\vect{\omega},n}(\overline{\delta \q})\|
    &\le \bm{\rho}^{(\vect{\omega}),\ddagger}_{n}(\overline{\delta \q},\gamma_{\vect{\omega}}),\\[1mm]
    \|\e_{\vect{v},n}(\delta \Q)\|^{2}
    &\le \big(
            \bm{\rho}^{(\vect{v}),\ddagger}_{n}(\delta \Q,\gamma_{\vect{v}})
         \big)^{2}.
\end{align}
Define the worst-case constants
\begin{align}
    c^{(\vect{\omega})}_n
    &:= \max_{\overline{\delta \q}\in Q^{(\vect{\omega})}_{\mathcal{X}}}
        \bm{\rho}^{(\vect{\omega}),\ddagger}_{n}(\overline{\delta \q},\gamma_{\vect{\omega}}),
        \label{eq:def_c_omega}\\[1mm]
    c^{(\vect{v})}_n
    &:= \frac{1}{2\alpha_n^{(\vect{v})}}
        \max_{\delta \Q\in Q^{(\vect{v})}_{\mathcal{X}}}
        \big(
            \bm{\rho}^{(\vect{v}),\ddagger}_{n}(\delta \Q,\gamma_{\vect{v}})
        \big)^{2}.
        \label{eq:def_c_v}
\end{align}
Then, on the event of probability at least $\gamma$, inequality
\eqref{eq:Vdot_pre_prob} implies $\dot{V}_n
    \le - \alpha_n^{(\vect{\omega})} |\delta q_0| \|\delta \q\|^{2}
        - \tfrac{1}{2}\alpha_n^{(\vect{v})} \|\delta \p\|^{2}
        + c^{(\vect{\omega})}_n + c^{(\vect{v})}_n$.

Let $\alpha_n := \tfrac{1}{2}\min\{\alpha_n^{(\vect{\omega})},\alpha_n^{(\vect{v})}\}$, so that $\alpha_n^{(\vect{\omega})} \ge 2\alpha_n$ and $\tfrac{1}{2}\alpha_n^{(\vect{v})} \ge \alpha_n$. Then
\begin{equation}\label{eq:Vdot_alpha_n}
    \dot{V}_n
    \le - \alpha_n\big(
            |\delta q_0| \|\delta \q\|^{2}
            + \|\delta \p\|^{2}
         \big)
        + d_n,
\end{equation}
where we define $d_n := c^{(\vect{\omega})}_n + c^{(\vect{v})}_n >0$. Introduce the scalar $\varepsilon_{0,n} := \frac{d_n}{\alpha_n}$, and the set
\begin{equation}
\mathcal{A}_{\varepsilon_{0,n}}
:= \left\{
      \delta \Q \;:\;
      |\delta q_0|\,\|\delta \q\|^{2}
      + \|\delta \p\|^{2}
      > \varepsilon_{0,n}
   \right\}.
\end{equation}
For all $\delta \Q\in\mathcal{A}_{\varepsilon_{0,n}}$ we have
\[
|\delta q_0| \|\delta \q\|^{2}
+ \|\delta \p\|^{2}
> \varepsilon_{0,n}
= \frac{d_n}{\alpha_n},
\]
and therefore, by \eqref{eq:Vdot_alpha_n}, $\dot{V}_n
< - \alpha_n \varepsilon_{0,n} + d_n
= 0$. Thus, conditioned on the event of probability at least $\gamma$ where the GP
error bounds hold, we obtain $\dot{V}_n(\delta \Q) < 0$ for all $\delta \Q\in\mathcal{A}_{\varepsilon_{0,n}}$.

Now consider the complementary set
\[
K_{\varepsilon_{0,n}}
:= \left\{
      \delta \Q \;:\;
      |\delta q_0|\,\|\delta \q\|^{2}
      + \|\delta \p\|^{2}
      \le \varepsilon_{0,n}
   \right\},
\]
as in the statement of the theorem, and define $M_n := \max_{\delta \Q\in K_{\varepsilon_{0,n}}} V_n(\delta \Q)$. Since $K_{\varepsilon_{0,n}}$ is compact and $V_n$ is continuous,
$M_n$ is finite and strictly positive (unless $\delta \Q$ is at the origin). Let $B_n := \{\delta \Q : V_n(\delta \Q)\le M_n\}$. Then $K_{\varepsilon_{0,n}}\subseteq B_n$. From \eqref{eq:Vdot_alpha_n}
and the negativity of $\dot{V}_n$ on $\mathcal{A}_{\varepsilon_{0,n}}$,
standard Lyapunov arguments imply that, conditioned on the event of probability
at least $\gamma$,
every trajectory $\delta \Q(t)$ enters $B_n$ in finite time and,
once inside $B_n$, it cannot leave $B_n$ (otherwise $\dot{V}_n$ would
have to be positive at some point on the boundary of $B_n$, which contradicts
\eqref{eq:Vdot_alpha_n}). Hence $B_n$ is a positively invariant set, and
$K_{\varepsilon_{0,n}}$ is uniformly ultimately attractive.

By Assumption~\ref{assumption:1}, the number of dataset switches $n(t)$ is
finite and there exists $n_{\mathrm{end}}$ such that $n(t)=n_{\mathrm{end}}$
for all $t\ge T_{\mathrm{end}}$. Using the multiple Lyapunov function
argument~\cite{liberzon1999basic}, and the fact that each $V_n$ decreases
outside the corresponding $K_{\varepsilon_{0,n}}$ with probability at least
$\gamma$, we conclude that there exists $T\ge 0$ such that, for all $t\ge T$, $V_{n_{\mathrm{end}}}(\delta \Q(t))
\le M_{n_{\mathrm{end}}}$ with probability at least $\gamma$. This is equivalent to
\[
\|\delta \q(t)\|^{2}
+ \tfrac{1}{2}\|\delta \p(t)\|^{2}
\le M_{n_{\mathrm{end}}},
\quad \forall t\ge T,
\]
with probability at least $\gamma$, which proves the uniform ultimate boundedness
in probability stated in \eqref{modelerror}.
\hfill$\square$

\begin{remark}
This framework supports \textit{online learning} through the sequential update of the datasets $\mathcal{D}_n^{(\vect{\omega})}$ and $\mathcal{D}_n^{(\vect{v})}$, which are enlarged or replaced as new data become available. 
In this work, however, the emphasis is placed on the stability analysis of the resulting learning-based controller rather than on the optimization of online GP performance. 
Advanced strategies for real-time GP updates, including sparse and streaming formulations, can be found in~\cite{dai2024cooperative,dai2024decentralized}.
\end{remark}

\begin{remark}
The Gaussian Process models $\mathrm{GP}^{(\vect{\omega})}$ and $\mathrm{GP}^{(\vect{v})}$ 
can also accommodate processing or communication delays by conditioning the posterior on delayed measurements. 
This mechanism enables posterior corrections of the estimation error when new or delayed data arrive, 
which enhances the robustness of the GP-based controller against feedback latency and asynchronous data acquisition.
\end{remark}

\subsection{Online learning-based tracking algorithm}

The implementation procedure naturally follows the theoretical framework established in 
Theorem~\ref{thm:lbc}. In particular, at each learning iteration $n$, the Gaussian Process models 
$\mathrm{GP}^{(\vect{\omega})}$ and $\mathrm{GP}^{(\vect{v})}$ are retrained with the updated datasets 
$\mathcal{D}_n^{(\vect{\omega})}$ and $\mathcal{D}_n^{(\vect{v})}$, yielding new posterior mean 
estimates $\widehat{\bm{\rho}}_{\vect{\omega},n}$ and $\widehat{\bm{\rho}}_{\vect{v},n}$. 
These estimates define the learned control laws~\eqref{learnedcontrol}–\eqref{learnedcontrolv} and 
the corresponding Lyapunov function $V_n(\delta\Q)$. 

Optionally, the probabilistic model-error bounds 
$\bm{\rho}^{(\vect{\omega}),\ddagger}_{n}$ and $\bm{\rho}^{(\vect{v}),\ddagger}_{n}$ 
can be computed using~\eqref{eq:rho_ddagger_compact} to evaluate the ultimate bound $M_n$ of~\eqref{modelerror}. 
In particular, the final iteration $n=n_{\text{end}}$ yields the bound $M_{n_{\text{end}}}$, which 
represents the maximum expected steady-state tracking error with confidence level 
$\gamma=\min\{\gamma_{\vect{\omega}},\gamma_{\vect{v}}\}$. 
This iterative procedure is summarized in Algorithm~\ref{alg:1}

\begin{figure}[h!]
\vspace{-0.2cm}
\begin{algorithm}[H]
\caption{Online learning-based tracking control}
\label{alg:1}
\begin{algorithmic}
\State \textbf{Input:} maximum number of GP updates $n_{\text{end}}$; batch sizes $\{m_n\}_{n=1}^{n_{\text{end}}}$; 
initial datasets $\mathcal{D}_0^{(\vect{\omega})},\mathcal{D}_0^{(\vect{v})}$ (possibly empty)
\State \textbf{Initialize:} train the GP models $f_{GP}^{(\vect{\omega})}$ and $f_{GP}^{(\vect{v})}$ on 
$\mathcal{D}_0^{(\vect{\omega})},\mathcal{D}_0^{(\vect{v})}$ and obtain posterior means 
$\widehat{\bm{\rho}}_{\vect{\omega},0},\widehat{\bm{\rho}}_{\vect{v},0}$
\State \textbf{Initialize:} implement the control laws $\widetilde{\vect{\omega}}_{c},\widetilde{\vect{v}}_{c}$ 
according to~\eqref{learnedcontrol}–\eqref{learnedcontrolv} with $n=0$
\State Set $n \gets 0$
\While{$n < n_{\text{end}}$}
    \State $n \gets n+1$
    \State Collect $m_n$ new data points from the system~\eqref{omegabar}–\eqref{vbar} and 
    update the datasets $\mathcal{D}_n^{(\vect{\omega})},\mathcal{D}_n^{(\vect{v})}$ as in~\eqref{eq:gp_dataset}
    \State Retrain the GP models on $\mathcal{D}_n^{(\vect{\omega})},\mathcal{D}_n^{(\vect{v})}$ and 
    compute the posterior means $\widehat{\bm{\rho}}_{\vect{\omega},n},\widehat{\bm{\rho}}_{\vect{v},n}$
    \State (Optional) compute the error bounds 
    $\bm{\rho}^{(\vect{\omega}),\ddagger}_{n}$ and $\bm{\rho}^{(\vect{v}),\ddagger}_{n}$ 
    using~\eqref{eq:rho_ddagger_compact} and the associated ultimate bound $M_n$ in~\eqref{modelerror}
    \State Update the control laws $\widetilde{\vect{\omega}}_{c},\widetilde{\vect{v}}_{c}$ 
    using the new posterior means in~\eqref{learnedcontrol}–\eqref{learnedcontrolv}
\EndWhile
\end{algorithmic}
\end{algorithm}
\vspace{-0.8cm}
\end{figure}

\medskip

At the beginning of the procedure, the designer sets the maximum allowable tracking error $M_{n_{\text{end}}}$ and the maximum number $n_{\text{end}}$ of dataset updates for the Gaussian Process models. 
This choice depends on the total amount of data that can be stored and processed, which is typically limited by memory and computational resources. 
If $m_n$ data points are collected between instants $t_n$ and $t_{n+1}$, and the total available storage capacity allows for $\bar{m}$ data points, then the maximum number of datasets is chosen as 
$n_{\text{end}} = \lfloor \bar{m}/m_n \rfloor$.

Each Gaussian Process model, $\mathrm{GP}^{(\vect{\omega})}$ and $\mathrm{GP}^{(\vect{v})}$, is initialized with its corresponding dataset $\mathcal{D}_0^{(\vect{\omega})}$ and $\mathcal{D}_0^{(\vect{v})}$. 
These initial datasets may come from an offline identification experiment, a simulation campaign, or, if no prior data are available, they can be empty. 
In the latter case, the initial GP posterior means are zero, i.e., 
$\widehat{\vect{\rho}}_{\vect{\omega},0} = \vect{0}$ and $\widehat{\vect{\rho}}_{\vect{v},0} = \vect{0}$, 
so the control laws reduce to their nominal form \eqref{omegabar}–\eqref{vbar}.

During operation, a batch of $m_n$ new measurements is collected from the vehicle between $t_n$ and $t_{n+1}$ and stored as datasets 
$\mathcal{D}_n^{(\vect{\omega})}$ and $\mathcal{D}_n^{(\vect{v})}$. 
Both Gaussian Process models are then retrained with their respective datasets, and at time $t_{n+1}$ the control inputs are updated according to the learned compensations 
$\widehat{\vect{\rho}}_{\vect{\omega},n}$ and $\widehat{\vect{\rho}}_{\vect{v},n}$ 
as given by \eqref{learnedcontrol}–\eqref{learnedcontrolv}.

Importantly, up to time $t_n$ the control loop continues to use the previous estimates 
$\widehat{\vect{\rho}}_{\vect{\omega},n-1}$ and $\widehat{\vect{\rho}}_{\vect{v},n-1}$, 
which means that there are no hard real-time constraints on the GP retraining step. 
This process is repeated until the maximum number of datasets $n_{\text{end}}$ is reached, 
after which the stability result of Theorem~\ref{thm:lbc} guarantees that the tracking error $\delta \Q(t)$ remains within the probabilistic bound defined by $M_{n_{\text{end}}}$ with confidence level $\gamma = \min\{\gamma_{\vect{\omega}},\gamma_{\vect{v}}\}$.

\begin{remark}
The proposed framework can be regarded as a geometric extension of the 
online learning-based balancing (OLBB) method~\cite{beckers2022learning} 
from Euclidean state spaces of second-order mechanical systems 
to the dual-quaternion representation of $\mathrm{SE}(3)$. 
In~\cite{beckers2022learning}, the OLBB controller combines a model-based term 
with an adaptive feedback gain law whose magnitude depends on the model error 
predicted by a Gaussian process oracle, ensuring bounded tracking error with high probability. 
In the present work, this principle is reformulated in geometric terms: 
the unknown rotational and translational dynamics, 
$\bm{\rho}_{\vect{\omega}}$ and $\bm{\rho}_{\vect{v}}$, 
play the role of the unmodeled forces in OLBB, 
and the probabilistic bounds derived from the GP posterior 
enter the Lyapunov analysis in Theorem~\ref{thm:lbc} 
to guarantee bounded dual-quaternion tracking error. 
Hence, Theorem~\ref{thm:lbc} generalizes the OLBB stability result 
to the nonlinear manifold of rigid-body motions, 
preserving the probabilistic learning guarantees while respecting 
the group structure of $\mathrm{SE}(3)$.
\end{remark}

\section{Simulation Results}\label{sec:exp}

To validate the proposed algorithm, several numerical simulations were carried out. The first simulation corresponds to a lemniscate-shaped trajectory, shown in Fig. \ref{fig:real_trajectory}. In this figure, the nominal reference path and the attitude that the vehicle is expected to follow are depicted.

The objective of this simulation is twofold. 
First, it illustrates the nominal
trajectory–tracking performance of the dual-quaternion controller under a
time-varying reference motion. Second, it evaluates the ability of the proposed
learning-based compensation scheme to handle persistent, state-dependent
disturbances that affect both rotational and translational dynamics.

\begin{figure}[h!]
\centering
\includegraphics[width=0.9\linewidth]{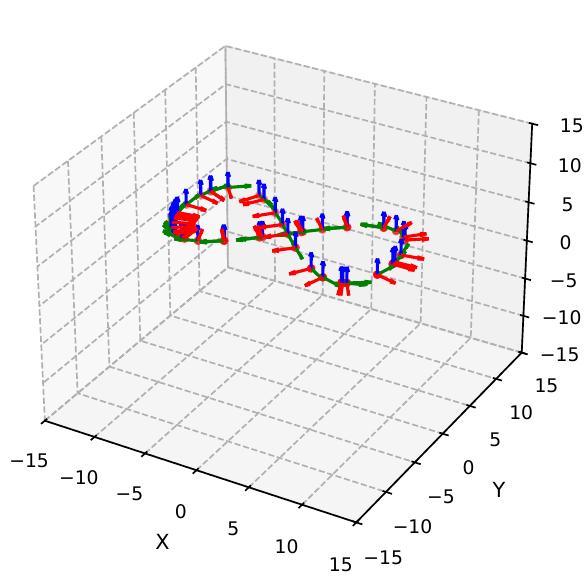}
\caption{Nominal reference trajectory for position and attitude.}
\label{fig:real_trajectory}
\end{figure}

The vehicle travels along this trajectory with a linearly decreasing velocity, meaning that as time progresses, the vehicle moves more slowly along the lemniscate. As a consequence, the vehicle spends an increasing amount of time near the center of the trajectory.

At the center of the lemniscate, an external disturbance source is assumed to be present. This disturbance affects both the vehicle orientation (yaw angle) and its vertical position (altitude). Therefore, whenever the vehicle approaches the center of the trajectory, an external perturbation acts on the system. The effect of this perturbation depends on how long the vehicle remains close to the center: as the vehicle slows down over time, the disturbance acts for a longer duration.

From a modeling perspective, this disturbance can be interpreted as a localized,
state-dependent perturbation acting on both the angular velocity and linear velocity channels. In particular, the disturbance is activated when the vehicle state approaches a specific region and remains inactive elsewhere. This setup is consistent with the disturbance model introduced in Section~\ref{sec:ps}, where unknown functions affect the rotational and translational
kinematics in a coupled manner.

This behavior can be clearly observed in Figs. \ref{fig:atti_perturbation} and \ref{fig:pos_perturbation}, where the duration of the perturbation increases as time evolves. This is a direct consequence of the decreasing vehicle speed, which causes the vehicle to remain longer in the vicinity of the disturbance source. Figures~\ref{fig:atti_perturbation} and~\ref{fig:pos_perturbation} report the injected
disturbances together with the corresponding GP-based estimates. The increasing duration of the perturbation windows reflects the decreasing vehicle speed, while the GP predictions adapt online to the repeated exposure to the same
disturbance pattern.

\begin{figure}[h!]
\centering
\includegraphics[width=0.9\linewidth]{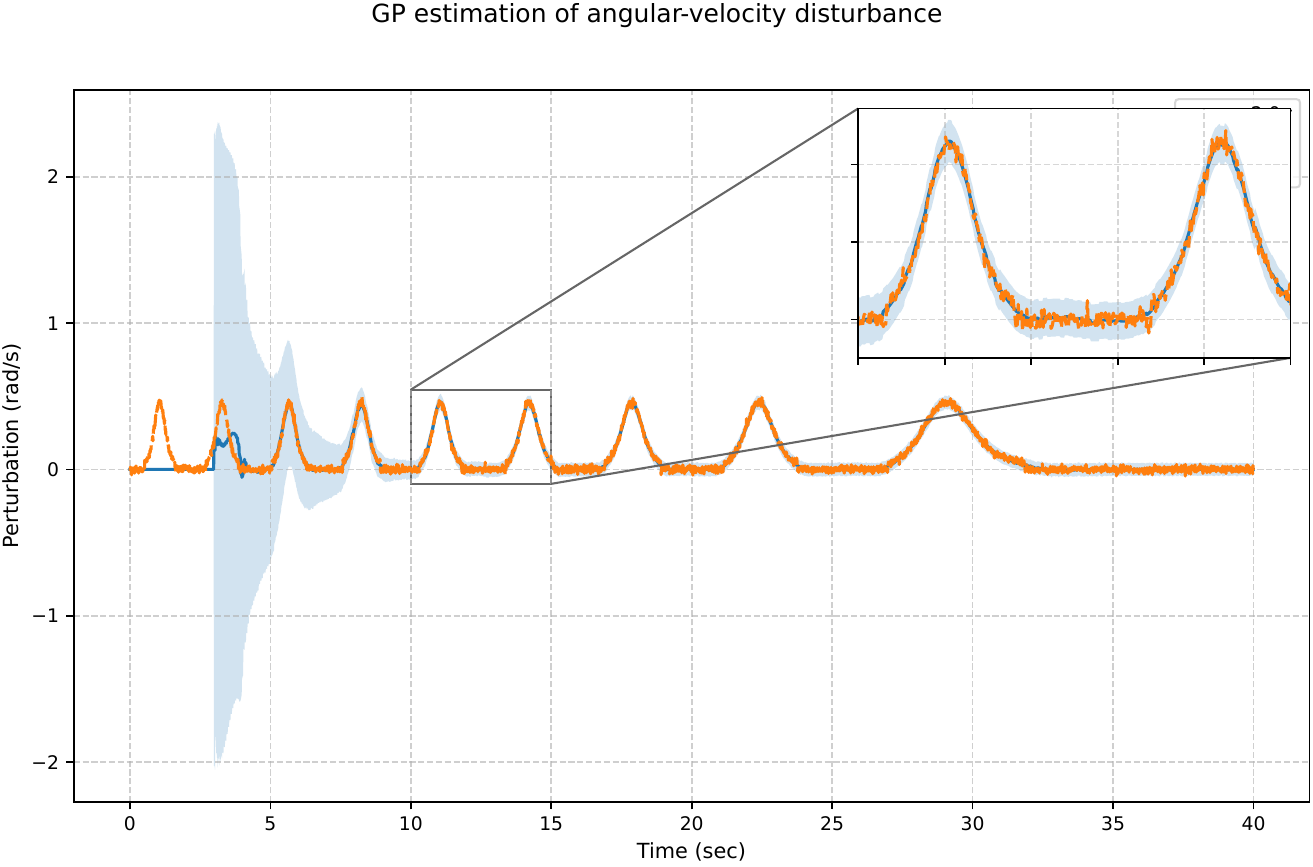}
\caption{External disturbance affecting yaw and altitude when the vehicle approaches the center of the trajectory. GP estimation of the angular-velocity disturbance.}
\label{fig:atti_perturbation}
\end{figure}

\begin{figure}[h!]
\centering
\includegraphics[width=0.9\linewidth]{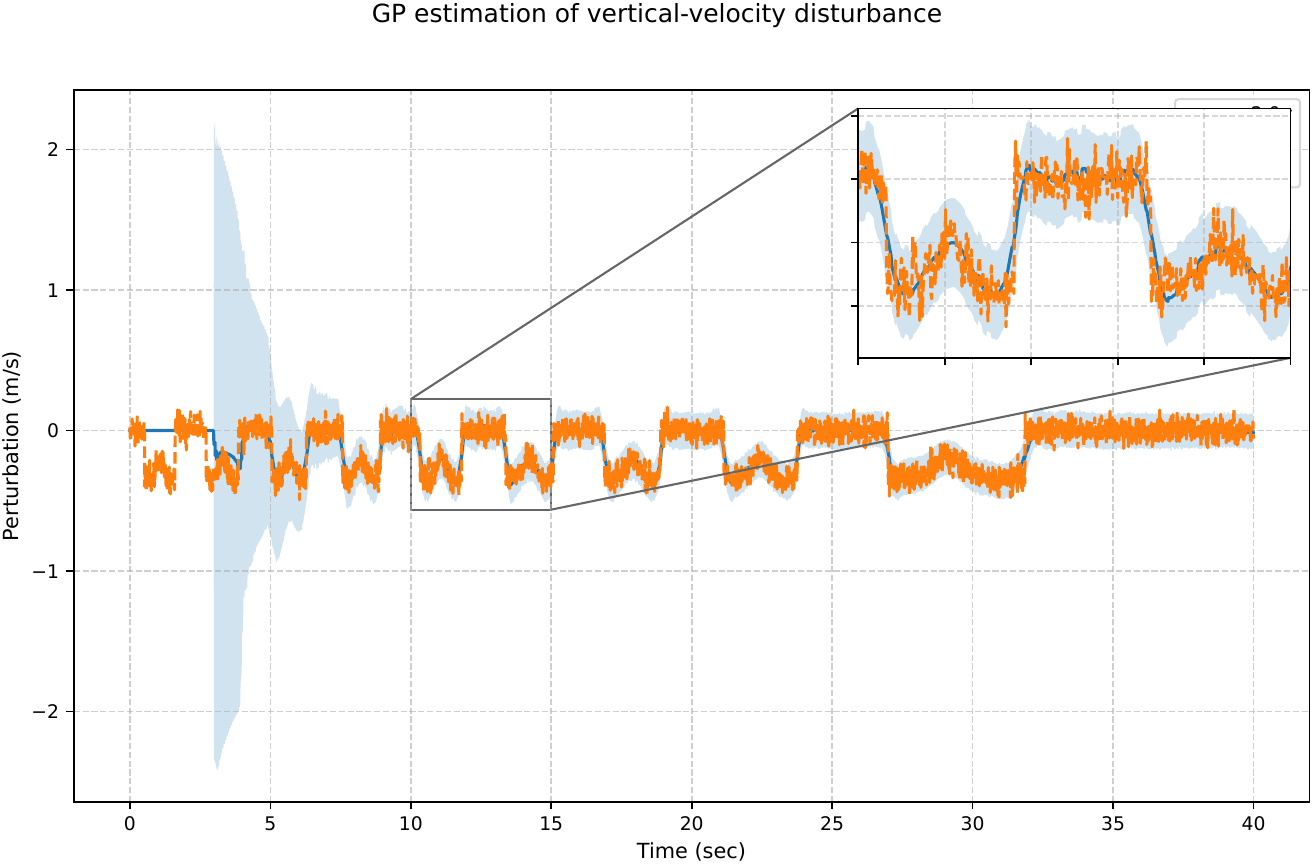}
\caption{External disturbance affecting yaw and altitude when the vehicle approaches the center of the trajectory. GP estimation of the linear velocity disturbance.}
\label{fig:pos_perturbation}
\end{figure}

From a physical perspective, this type of perturbation may represent an external magnetic disturbance affecting the magnetometer measurements. Such a disturbance causes a deflection in the yaw angle, which forces the vehicle to adjust the rotational speeds of pairs of rotors (clockwise or counter-clockwise) to compensate for the undesired rotation. Due to imperfections in the vehicle dynamic model—particularly in the matrix that maps torque and thrust to motor speeds—these rotor speed changes can temporarily reduce the total thrust. As a result, a loss of altitude occurs, which becomes correlated with the magnetic disturbance affecting the yaw measurement. This choice deliberately increases the difficulty of the tracking task, since the
vehicle is exposed to the disturbance for progressively longer time intervals.
As a result, the accumulated effect of the perturbation grows over time, providing
a stringent test for the learning-based compensation mechanism. 

\subsection{Open-Loop Experiment (Without GP Compensation)}

Two numerical experiments were conducted under identical conditions. In the first experiment, the controller does not compensate the disturbances using the Gaussian Process (GP) estimates (i.e., the GP runs online but its predictions are not injected into
the control inputs). For clarity, this is an ``open-loop'' case only with respect to the learning
compensation; the feedback controller itself remains closed-loop.

As shown in Fig. \ref{fig:atti_trackin_ol}, when the magnetic field sensed by the magnetometer is affected by the disturbance, the yaw tracking performance degrades noticeably. At the same time, Fig. \ref{fig:pos_trackin_ol} shows that the position tracking is also affected, with the disturbance being particularly evident in the loss of altitude.

\begin{figure}[t]
\centering
\includegraphics[width=0.9\linewidth]{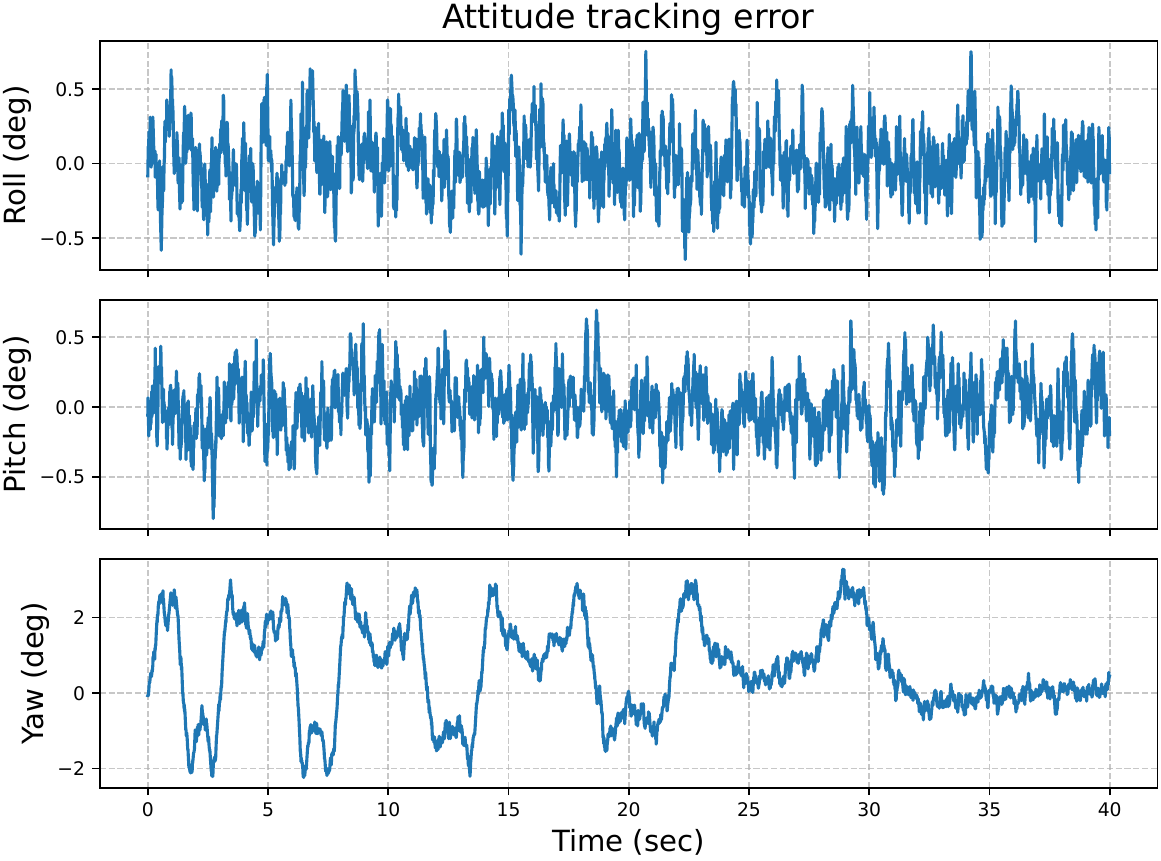}
\caption{Open-loop yaw tracking performance in the presence of external magnetic disturbances.}
\label{fig:atti_trackin_ol}
\end{figure}

\begin{figure}[t]
\centering
\includegraphics[width=0.9\linewidth]{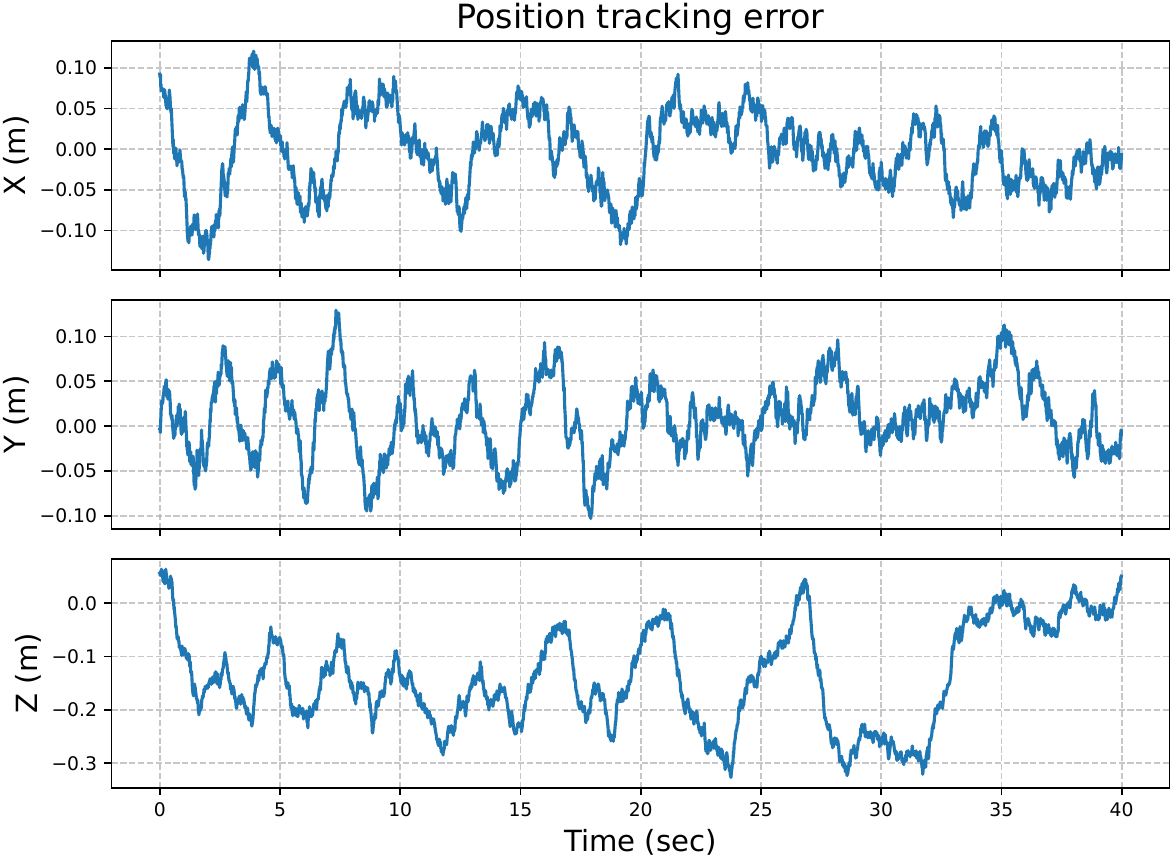}
\caption{Open-loop position tracking performance, showing altitude loss induced by the external disturbance.}
\label{fig:pos_trackin_ol}
\end{figure}

To quantitatively assess the controller performance, the Mean Squared Error (MSE) and Mean Absolute Error (MAE) were computed. In order to analyze how these metrics evolve over time, the errors were evaluated over sliding time windows of 10 s. This approach allows observing the temporal evolution of the tracking performance. The resulting MSE and MAE for attitude and position are shown in Figs. \ref{fig:mse_att_ol} and \ref{fig:mse_pos_ol}, respectively.

\begin{figure}[t]
\centering
\includegraphics[width=0.9\linewidth]{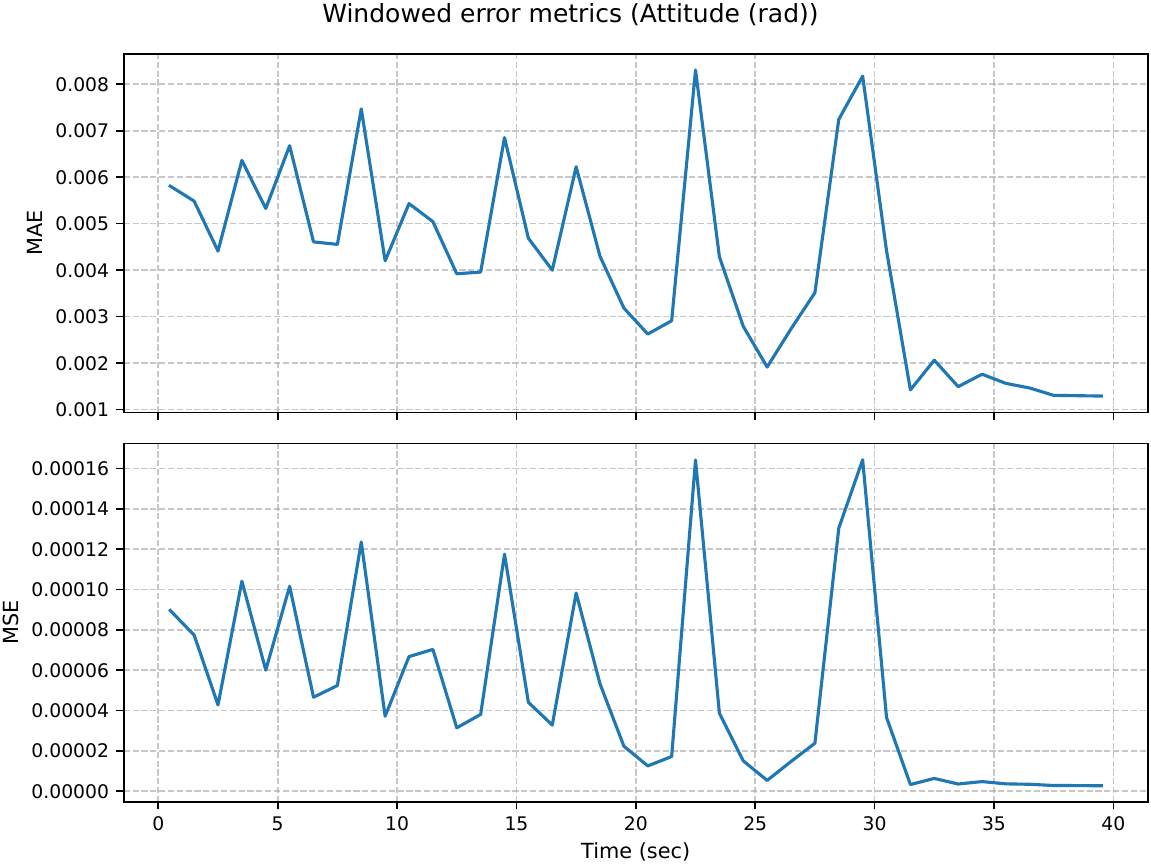}
\caption{Mean Absolute Error (MAE) and Mean Squared Error (MSE) of the attitude tracking error, computed over sliding time windows of 10~s, for the open-loop controller.}
\label{fig:mse_att_ol}
\end{figure}

\begin{figure}[t]
\centering
\includegraphics[width=0.9\linewidth]{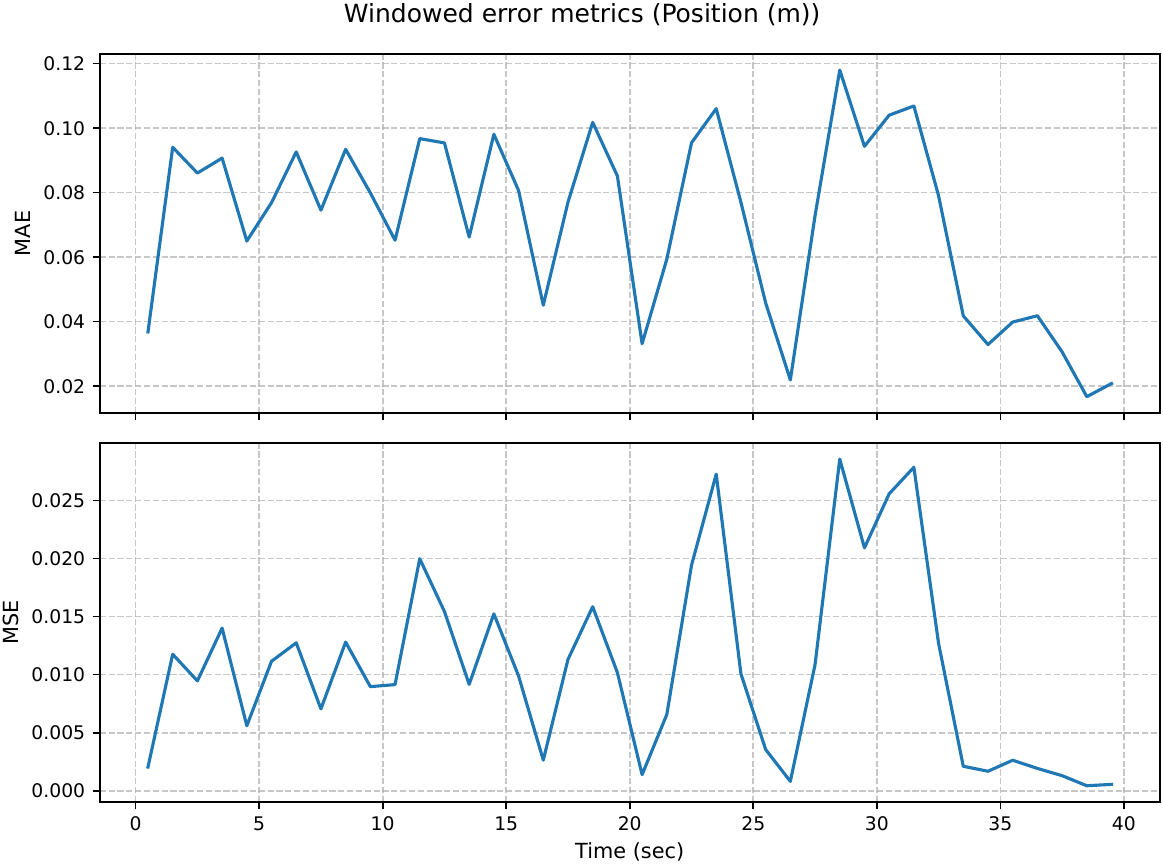}
\caption{Mean Absolute Error (MAE) and Mean Squared Error (MSE) of the position tracking error, computed over sliding time windows of 10~s, for the open-loop controller.}
\label{fig:mse_pos_ol}
\end{figure}

Although the GP estimates are not used for control compensation in this experiment, the GP estimator is still running online. This allows evaluating the estimation capability independently from the control loop. As shown in Figs. \ref{fig:atti_perturbation} and \ref{fig:pos_perturbation}, the GP provides accurate estimates of the disturbances affecting the vehicle. In these figures, the true disturbance is shown as a dashed orange line, the estimated mean as a solid blue line, and $\pm2$ standard deviations as a blue shaded region.

At the beginning of the simulation, the GP exhibits a large variance due to the lack of sufficient training data. However, as more samples are collected over time, the variance progressively decreases, and the GP converges to an accurate estimate of the external disturbances.

In these simulations, the navigation system integrates IMU data at \(300\,\text{Hz}\) and position and attitude information (from a magnetometer) at \(5\,\text{Hz}\). The magnetometer measurements are corrupted by additive small-angle noise (approximately \(0.5^\circ\)), while the position measurements are affected by zero-mean Gaussian noise with a standard deviation of \(0.5\,\text{m}\) per axis. Angular velocity is provided by a gyroscope whose noise characteristics are consistent with those of a MEMS IMU commonly used in low-cost flight controllers, with an Angle Random Walk (ARW) of approximately \(1.0\,\text{deg}/\sqrt{\text{h}}\). This information is fused using a standard Extended Kalman Filter (EKF) to estimate the vehicle’s position, velocity, and attitude. These estimates are then provided to the control algorithm, which runs at \(100\,\text{Hz}\).

The disturbance estimation is performed using Sparse Variational Gaussian Processes (SVGPs) implemented with the GPyTorch library. Two independent GP models are employed: one to estimate the disturbance affecting the attitude dynamics and another to estimate the disturbance affecting the position dynamics.

For both models, the kernel of Remmark \ref{remark2} with Automatic Relevance Determination (ARD) is used.

Training of the GP models is performed online in a sliding-window fashion using the Adam optimizer with a learning rate of \(10^{-2}\). 
A dataset of the most recent \(N = 2000\) samples is maintained, with an initial warm-up phase of 300 samples. Mini-batch of size of \(B = 256\) are used, and each GP update consists of five gradient-based optimization steps. This configuration enables the GP models to adapt online to slowly varying distrubances.

The proposed control and estimation framework is designed for real-time execution and was evaluated in terms of its computational requirements on embedded hardware. When deployed on an NVIDIA Jetson Orin Nano, the algorithm can be executed online thanks to the available GPU acceleration for both inference and online training of the SVGP models. In the considered implementation, the control loop operates at $100Hz$, while the GP models are updated at a lower rate of $20Hz$ using a sparse variational formulation with $128$ inducing points. 

Under these conditions, the execution time of the GP inference remains well below the control sampling period, and the periodic GP updates can be completed within a few milliseconds, allowing the controller to run in real time without violating timing constraints.

If additional computational margin is required, the control loop frequency can be reduced to values on the order of $50-60Hz$ without significantly affecting the control performance for the considered trajectories. This reduction directly relaxes the real-time constraints and further increases the available computation time for the GP updates, enabling either more frequent training steps or an increased number of optimization iterations. As a result, the algorithm remains suitable for real-time execution on embedded platforms while preserving the benefits of GP-based disturbance compensation.

\subsection{Close-Loop Experiment (With GP Compensation)}

In the second numerical experiment, the setup is identical to the first one, with the key difference that the control inputs are compensated using the GP disturbance estimates.

The resulting attitude and position tracking are shown in Figs. \ref{fig:atti_trackin_cl} and \ref{fig:pos_trackin_cl}. Initially, the compensation is not effective due to the high uncertainty of the GP estimates. However, once the GP has collected sufficient data, the compensation significantly improves the tracking performance.

\begin{figure}[t]
\centering
\includegraphics[width=0.9\linewidth]{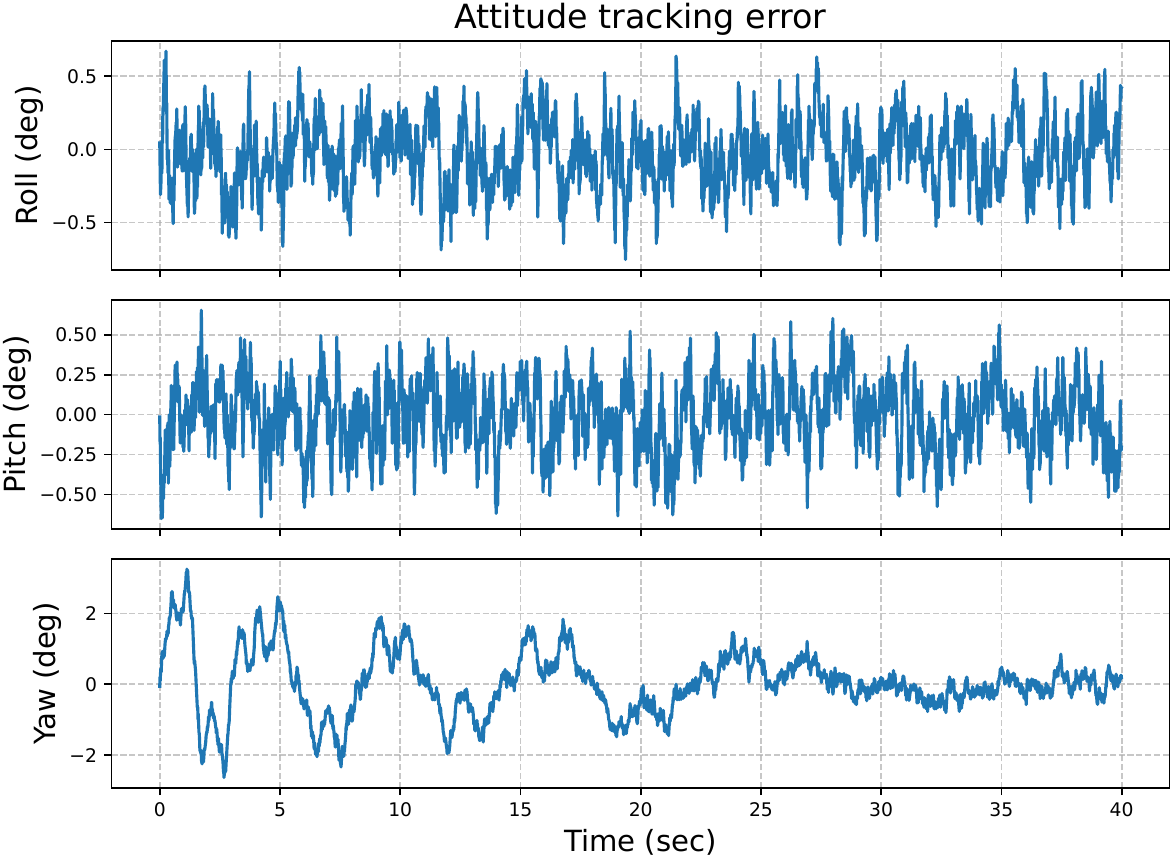}
\caption{Closed-loop attitude tracking performance with GP-based disturbance compensation. After an initial transient with high GP uncertainty, the yaw tracking accuracy improves significantly.}
\label{fig:atti_trackin_cl}
\end{figure}

\begin{figure}[t]
\centering
\includegraphics[width=0.9\linewidth]{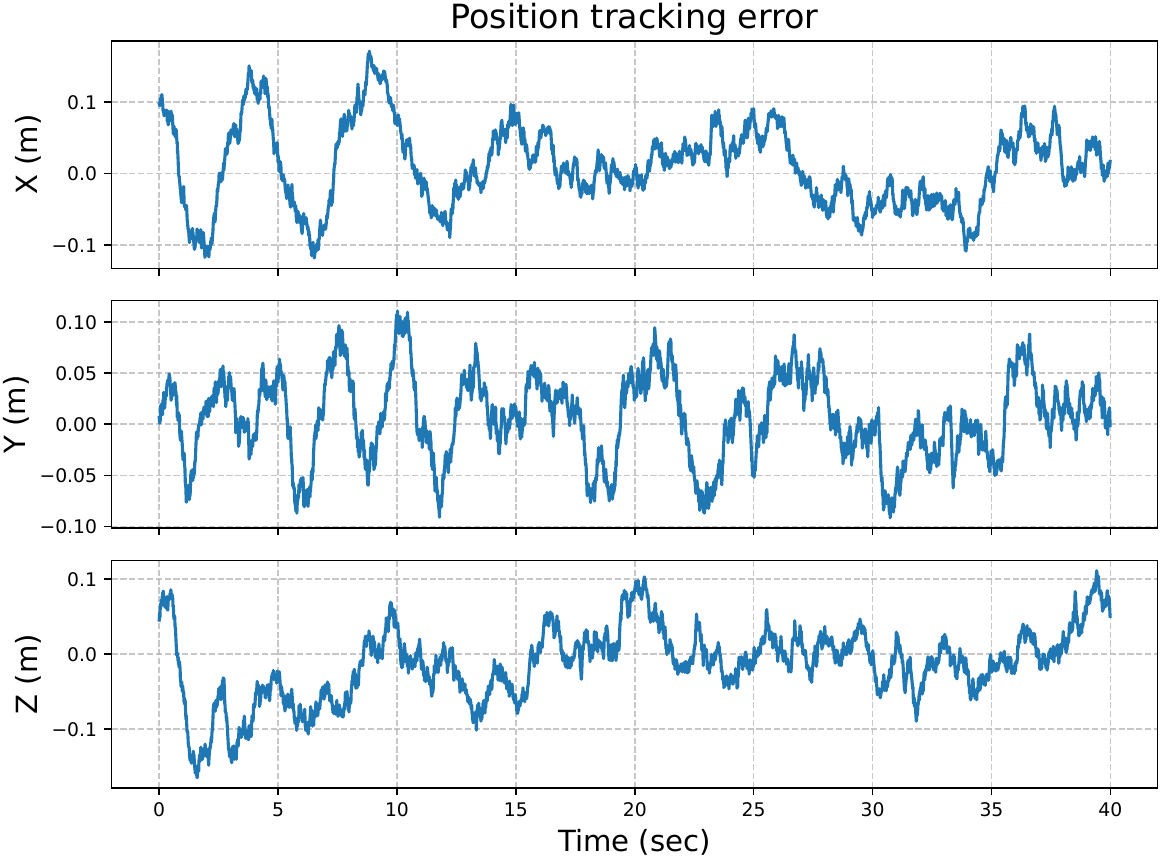}
\caption{Closed-loop position tracking performance with GP-based disturbance compensation. Once the GP estimates converge, the altitude loss induced by the disturbance is effectively mitigated.}
\label{fig:pos_trackin_cl}
\end{figure}

This improvement is quantitatively confirmed by the comparison of the MSE and MAE metrics shown in Figs. \ref{fig:mse_att_cl} and \ref{fig:mse_pos_cl}. When the GP-based compensation is enabled, both attitude and position errors are substantially reduced compared to the open-loop case.

\begin{figure}[t]
\centering
\includegraphics[width=0.9\linewidth]{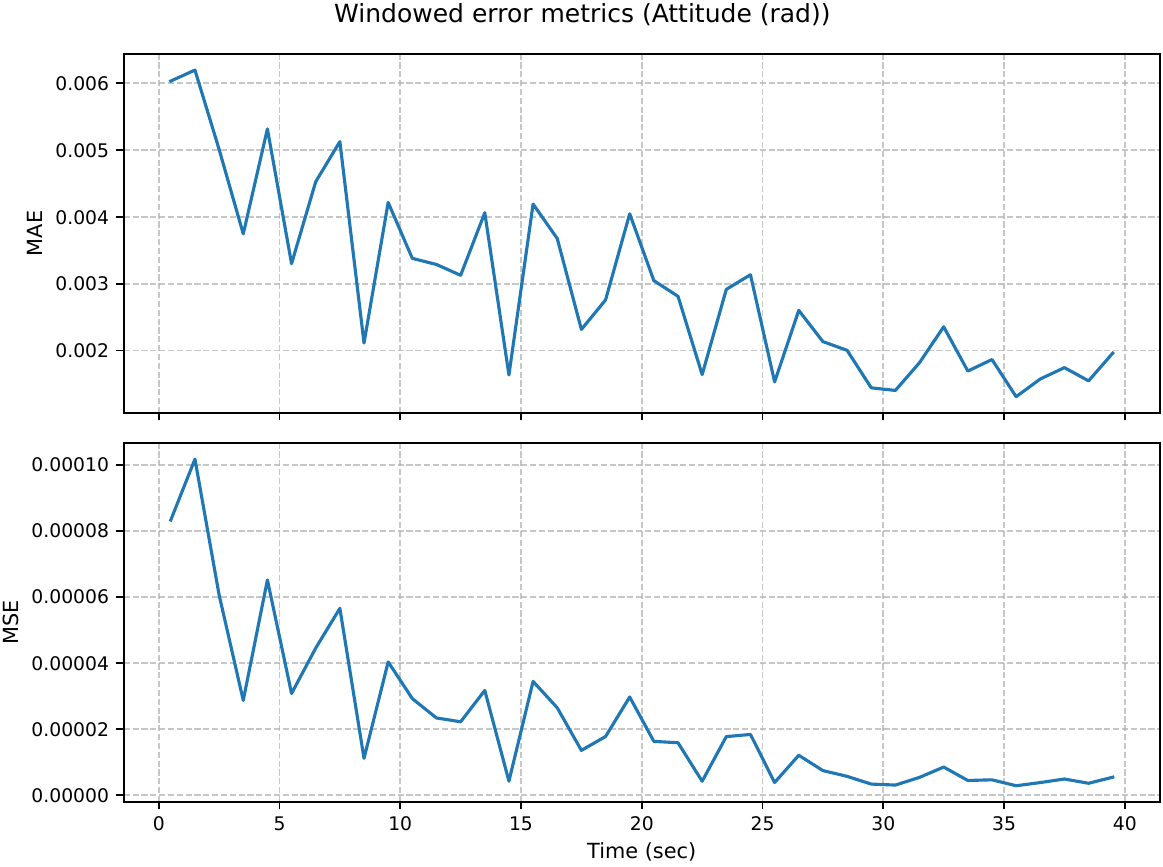}
\caption{Mean Absolute Error (MAE) and Mean Squared Error (MSE) of the attitude tracking error, computed over sliding time windows of 10~s, for the closed-loop controller with GP-based disturbance compensation.}
\label{fig:mse_att_cl}
\end{figure}

\begin{figure}[t]
\centering
\includegraphics[width=0.9\linewidth]{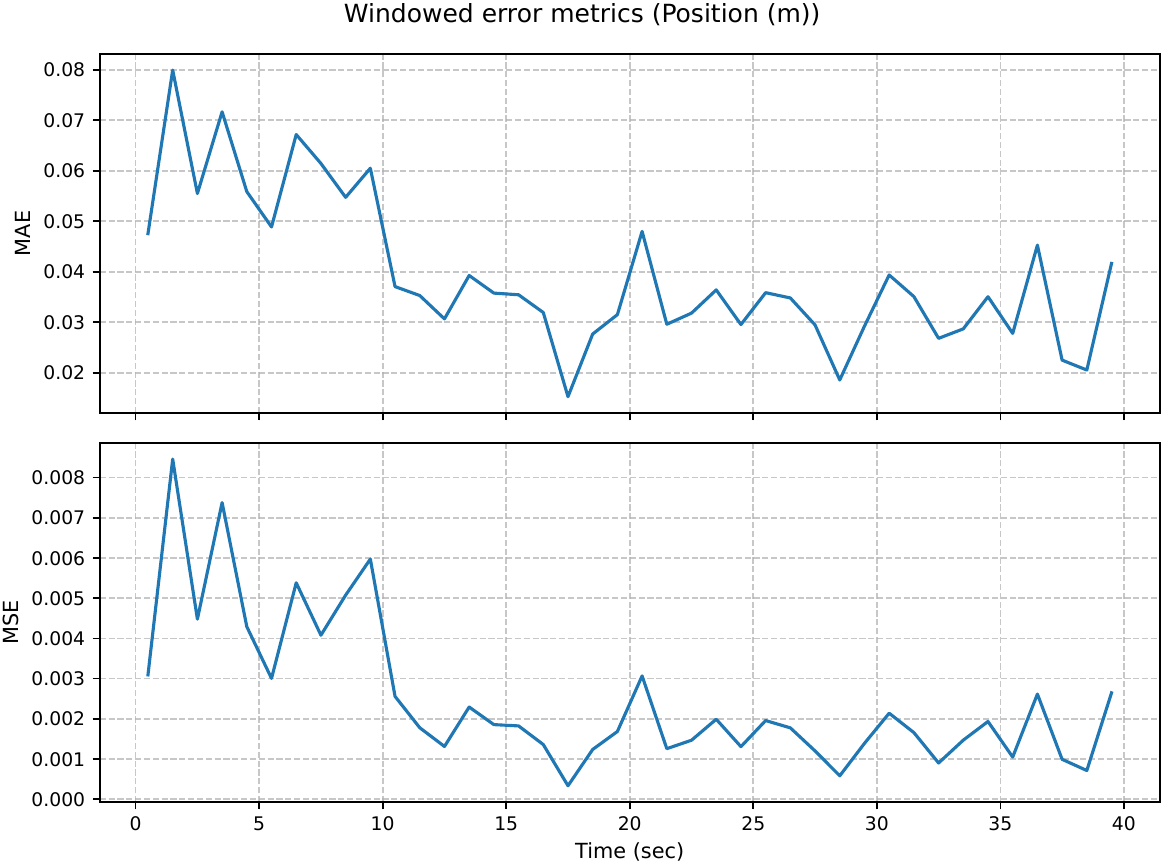}
\caption{Mean Absolute Error (MAE) and Mean Squared Error (MSE) of the position tracking error, computed over sliding time windows of 10~s, for the closed-loop controller with GP-based disturbance compensation.}
\label{fig:mse_pos_cl}
\end{figure}

\subsection{Quantitative Performance Evaluation on Different Trajectories}

To further analyze the behavior of the proposed algorithm, several numerical experiments were conducted using different reference trajectories. In particular, three representative classes of trajectories were considered: a lemniscate trajectory, a circular trajectory, and an ascending spiral trajectory. These $40s$-duration trajectories  were selected to evaluate the controller performance under different motion patterns and dynamic conditions. In these scenarios, the perturbation affects the vehicle throughout the entire flight.

For each class of trajectories $p=16$ experiments were performed, and the tracking performance was quantitatively assessed by computing the Mean Squared Error (MSE) and the Mean Absolute Error (MAE) of the attitude and position errors. The evaluation was performed both without and with the inclusion of the Gaussian Process (GP)-based disturbance compensation in the control loop.

The results demonstrate that incorporating the GP corrections significantly improves the tracking performance across all tested trajectories. This improvement is consistently observed in both attitude and position errors. A summary of the obtained MSE and MAE values for each trajectory, with and without GP-based compensation, is reported in Table~\ref{tab:performance_comparison}.

\begin{table}[t]
\centering
\caption{Mean Absolute Error (MAE) and Mean Squared Error (MSE) for different trajectories, with and without GP-based disturbance compensation.}
\label{tab:performance_comparison}
\begin{tabular}{lcc}
\hline
\, & \textbf{Attitude} & \, \\
\hline
\textbf{Trajectory} &
\textbf{MAE (GP)} &
\textbf{MAE (w/out GP)} \\
\hline
Lemniscate        & 0.00387 & 0.01950 \\
Circle            & 0.00441 & 0.02101 \\
Ascending spiral  & 0.00453 & 0.02123 \\
\hline
\end{tabular}
\begin{tabular}{lcc}
\hline
\textbf{Trajectory} &
\textbf{MSE (GP)} &
\textbf{MSE (w/out GP)} \\
\hline
Lemniscate        & 9.919e-05 & 0.00104 \\
Circle            & 0.00012 & 0.00121 \\
Ascending spiral  & 0.00012 & 0.00123 \\
\hline
\end{tabular}\\
\begin{tabular}{lcc}
\hline
\, & \textbf{Position} & \, \\
\hline
\textbf{Trajectory} &
\textbf{MAE (GP)} &
\textbf{MAE (w/out GP)} \\
\hline
Lemniscate        & 0.03752 & 0.11191 \\
Circle            & 0.04678 & 0.11775 \\
Ascending spiral  & 0.04579 & 0.12915 \\
\hline
\end{tabular}
\begin{tabular}{lcc}
\hline
\textbf{Trajectory} &
\textbf{MSE (GP)} &
\textbf{MSE (w/out GP)} \\
\hline
Lemniscate        & 0.00289 & 0.01927 \\
Circle            & 0.00378 & 0.01947 \\
Ascending spiral  & 0.00355 & 0.02357 \\
\hline
\end{tabular}
\end{table}

\section{Conclusions and Future Work}

This paper presented a learning-based trajectory tracking controller for autonomous
vehicles formulated in the dual quaternion framework.
By integrating Gaussian Process (GP) models into a geometric control law,
the proposed approach compensates unknown nonlinear dynamics and external disturbances
affecting both attitude and position, while preserving the geometric configuration of the vehicle. 
The probabilistic learning formulation enables data-driven adaptation without requiring
explicit parametric models of the uncertainties, and a Lyapunov-based analysis established
probabilistic ultimate boundedness of the pose tracking error under bounded GP uncertainty.

The results reported in Section~\ref{sec:exp} target a disturbance pattern that is
highly relevant for aerial robotics: localized magnetic anomalies that corrupt magnetometer readings and
primarily manifest as yaw errors. In the open-loop case, this yaw degradation propagates through the
attitude–thrust allocation pipeline: corrective rotor-speed differentials (needed to reject the spurious yaw)
interact with imperfect torque/thrust-to-motor mixing, producing a transient loss of net thrust and therefore
an altitude drop. This mechanism yields a practically meaningful coupling between rotational and translational
channels, clearly reflected by the simultaneous yaw and vertical-position degradations (Figs.~5--8).
When GP-based compensation is enabled, the controller learns this coupled, state-dependent effect directly
from flight data: the GP posterior quickly adapts to the repeated exposure to the disturbance region
(Figs.~3--4), and once uncertainty decreases the closed-loop tracking improves markedly in both yaw and
altitude (Figs.~9--12). Overall, these results highlight that the proposed geometric learning-based controller
can identify and compensate sensor-induced, cross-coupled effects in dual quaternion motion without requiring an
explicit parametric model of the underlying coupling, which is a common challenge in real-world deployments.

Future work will focus on extending the proposed framework to multi-agent systems
evolving on Lie groups.
A natural direction is the incorporation of motion feasibility conditions and
coordination constraints in learning-based controllers, building on recent results
for multi-agent systems on Lie groups~\cite{colombo2019motion}.
In this context, the integration of Gaussian Process learning with distributed and
consensus-based architectures may enable agents to share local models of uncertainty
and improve collective robustness, as explored in learning-based formation control
approaches~\cite{beckers2021onlineb, 11320442}.

Another promising research avenue is the extension of the dual-quaternion learning
framework to formation control and leader--follower architectures.
In particular, cluster-space formulations provide a powerful tool to decouple
intra-group shape variables from inter-group motion, enabling scalable coordination
strategies on dual quaternions~\cite{giribet2021dual}.
Embedding learning-based disturbance compensation within such cluster-space and
leader--follower structures would allow groups of heterogeneous vehicles to adapt
online to uncertain interactions while preserving geometric consistency.

Finally, future work will address large-scale experimental validation and real-time
implementations, including adaptive data selection, sparse GP approximations, and
heterogeneous vehicle dynamics.
These developments are expected to further enhance the scalability and applicability
of learning-based geometric control methods for cooperative robotic systems operating
in complex and uncertain environments.

\section*{Appendix}

\subsection{Relation between $\delta\Q$ and $\delta\Q_\rho$}

Recall that the pose of the vehicle is represented by the unit dual quaternion
\[
\Q = \quat{q} + \varepsilon\,\frac{1}{2}\,\tilde{p}\circ\quat{q},
\]
and the dual quaternion tracking error is defined as
\[
\delta\Q := \Q_d^*\circ\Q
= \overline{\delta\q} + \varepsilon\,\frac{1}{2}\,\widetilde{\delta p}_b\circ\overline{\delta\q},
\]
where $\overline{\delta\q} := \quat{q}_d^*\circ\quat{q}$ and 
$\widetilde{\delta p}_b = (\delta p_b,0)$ denotes the position error expressed in the desired body frame.

The sensor is affected by an unknown pose error modeled as the unit dual quaternion
\[
\Q_\rho := \quat{q}_\rho + \varepsilon\,\frac{1}{2}\,\tilde{p}_\rho\circ\quat{q}_\rho,
\]
where $\quat{q}_\rho$ represents the attitude error and $\tilde{p}_\rho = (p_\rho,0)$ a possible translational bias. 
The measured pose is then
\[
\Q_{\text{meas}} = \Q\circ\Q_\rho,
\]
and the corresponding dual quaternion error used in feedback is
\[
\delta\Q_\rho := \Q_d^*\circ\Q_{\text{meas}}
= \Q_d^*\circ\Q\circ\Q_\rho
= \delta\Q\circ\Q_\rho.
\]

Using the dual quaternion product
\[
(a + \varepsilon b)\circ(c + \varepsilon d) \;=\; a\circ c + \varepsilon\,(a\circ d + b\circ c),
\]
and substituting
\[
\delta\Q = \overline{\delta\q} + \varepsilon\,\frac{1}{2}\,\widetilde{\delta p}_b\circ\overline{\delta\q},\qquad
\Q_\rho = \quat{q}_\rho + \varepsilon\,\frac{1}{2}\,\tilde{p}_\rho\circ\quat{q}_\rho,
\]
we obtain
\begin{align*}
\delta\Q_\rho 
&= \bigl(\overline{\delta\q} + \varepsilon\,\tfrac{1}{2}\,\widetilde{\delta p}_b\circ\overline{\delta\q}\bigr)
   \circ\bigl(\quat{q}_\rho + \varepsilon\,\tfrac{1}{2}\,\tilde{p}_\rho\circ\quat{q}_\rho\bigr)\\[2pt]
&= \overline{\delta\q}\circ\quat{q}_\rho
   + \varepsilon\,\frac{1}{2}\Bigl(
      \overline{\delta\q}\circ\tilde{p}_\rho\circ\quat{q}_\rho 
      + \widetilde{\delta p}_b\circ\overline{\delta\q}\circ\quat{q}_\rho
     \Bigr).
\end{align*}

Hence, the principal and dual parts of $\delta\Q_\rho$ are
\begin{align}
\PrincipalPart(\delta\Q_\rho) 
&= \overline{\delta\q}_\rho 
:= \overline{\delta\q}\circ\quat{q}_\rho, \label{eq:deltaq_rho_principal}\\
\DualPart(\delta\Q_\rho) 
&= \frac{1}{2}\Bigl(
      \overline{\delta\q}\circ\tilde{p}_\rho\circ\quat{q}_\rho 
      + \widetilde{\delta p}_b\circ\overline{\delta\q}\circ\quat{q}_\rho
    \Bigr). \label{eq:deltaQ_rho_dual}
\end{align}

By Lemma~1, there exists a unique $\widetilde{\delta p}_b^\rho$ such that
\[
\delta\Q_\rho 
= \overline{\delta\q}_\rho + \varepsilon\,\frac{1}{2}\,\widetilde{\delta p}_b^\rho\circ\overline{\delta\q}_\rho,
\]
and comparing with \eqref{eq:deltaQ_rho_dual} we obtain
\begin{equation}
\widetilde{\delta p}_b^\rho 
= \Bigl(
      \overline{\delta\q}\circ\tilde{p}_\rho\circ\quat{q}_\rho 
      + \widetilde{\delta p}_b\circ\overline{\delta\q}\circ\quat{q}_\rho
  \Bigr)\circ\quat{q}_\rho^*\circ\overline{\delta\q}^*. 
\label{eq:deltap_rho_expression}
\end{equation}

In particular, if the translational bias vanishes, i.e. $\tilde{p}_\rho = 0$, then
\[
\DualPart(\delta\Q_\rho) 
= \frac{1}{2}\,\widetilde{\delta p}_b\circ\overline{\delta\q}\circ\quat{q}_\rho,
\qquad
\widetilde{\delta p}_b^\rho 
= \widetilde{\delta p}_b,
\]
and only the attitude part of the error is affected.

\bibliographystyle{IEEEtran}
\bibliography{cluster}

\begin{IEEEbiography}[{\includegraphics[width=1in,height=1.25in,clip,keepaspectratio]{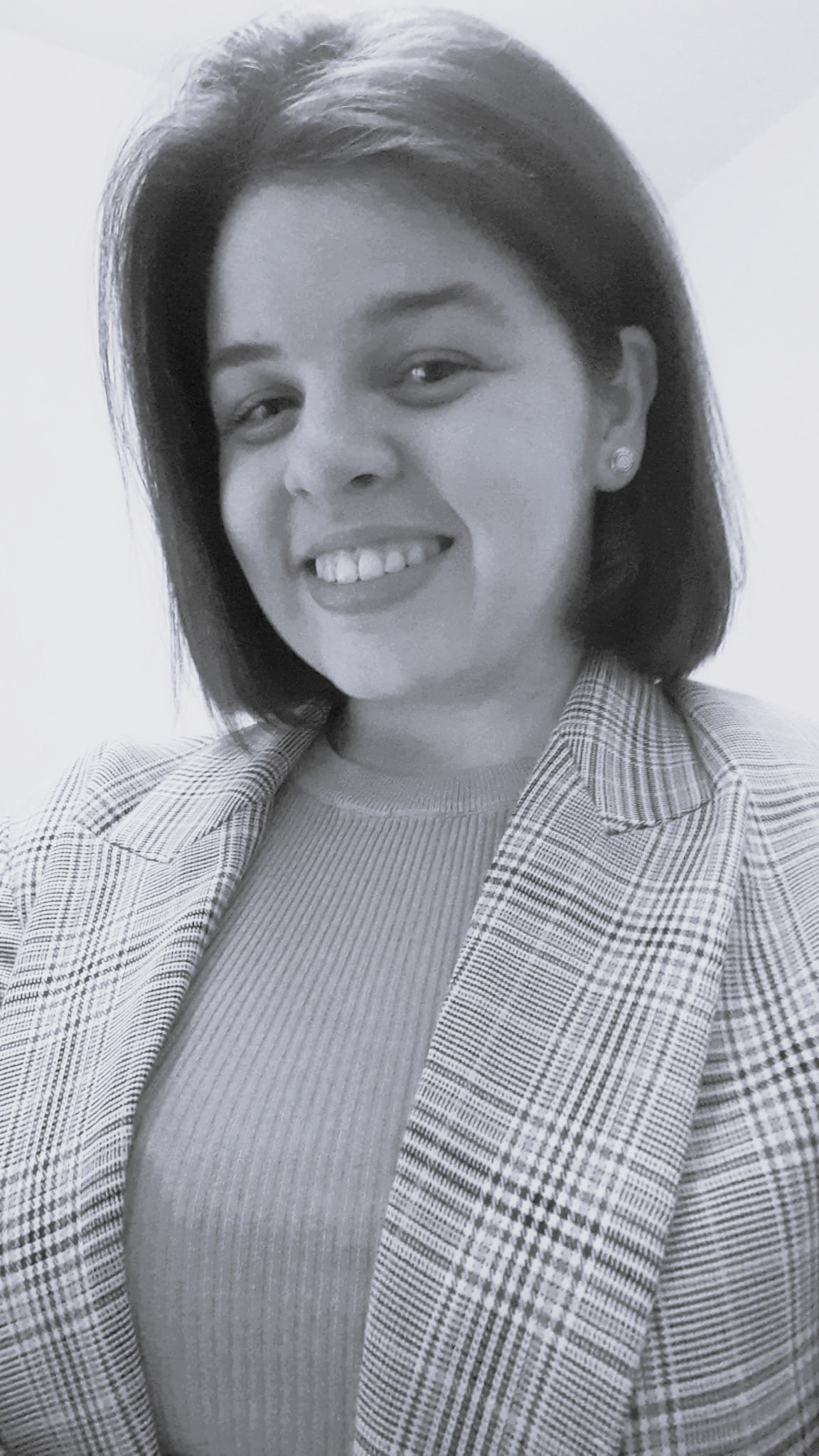}}]{Omayra Yago Nieto}
is a Ph.D. student  in Automation Control and Robotics at Universidad Politécnica de Madrid, Spain. Her research interests include geometric mechanics, learning-based control and machine learning methods.
\end{IEEEbiography}
\begin{IEEEbiography}[{\includegraphics[width=1in,height=1.25in,clip,keepaspectratio]{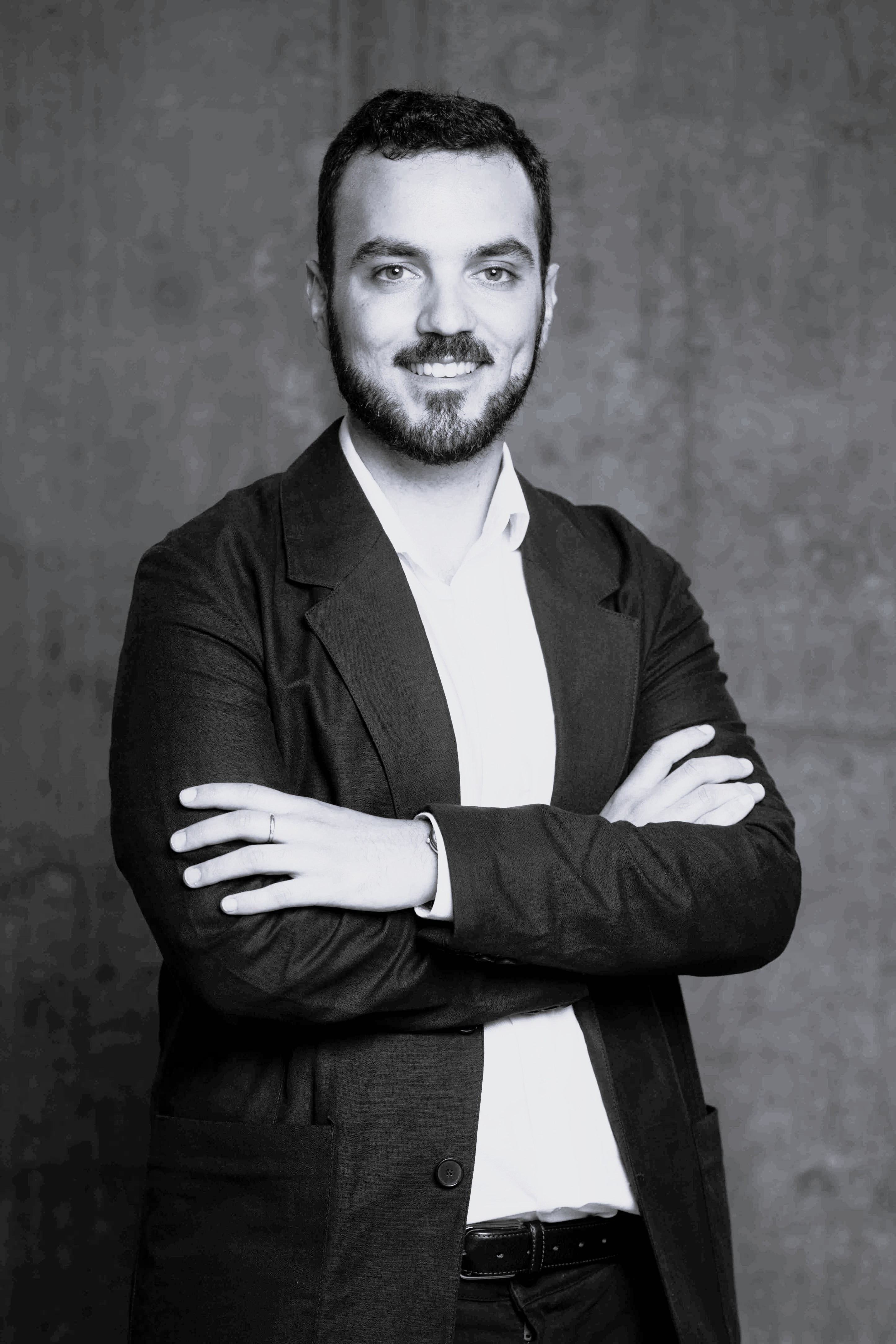}}]{Alexandre Anahory Simoes}
received a Ph.D. degree in Mathematics with Universidad Autonoma de Madrid, Spain and he is currently an Assistant Professor at IE School of Science and Technology in Madrid, Spain. His research interests include geometric mechanics, geometric control theory and numerical integration of dynamical systems.
\end{IEEEbiography}
\begin{IEEEbiography}[{\includegraphics[width=1in,height=1.25in,clip,keepaspectratio]{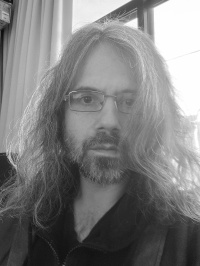}}]{Juan I. Giribet}
received the Electrical Engineering degree from the University of Buenos Aires, Argentina, in 2003, and the Ph.D. degree from the same institution in 2009. He is currently an Associate Professor at the University of San Andrés, where he is affiliated with the Artificial Intelligence and Robotics Laboratory, and a Researcher at CONICET in the area of applied mathematics. Prof. Giribet is a Senior Member of the IEEE.
\end{IEEEbiography}
\begin{IEEEbiography}[{\includegraphics[width=1in,height=1.25in,clip,keepaspectratio]{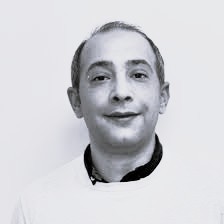}}]{Leonardo J. Colombo}
is a Tenured Scientist (Científico Titular) of the Spanish National Research Council (CSIC) at the Centre for Automation and Robotics (CAR). He holds a Ph.D. in Mathematics from the Autonomous University of Madrid (2014). In the period between 2014 and 2017, he was an Assistant Professor and Postdoctoral Researcher in the Department of Mathematics of the University of Michigan. He continued his postdoctoral training (2017-2018) as an ACCESS Lineaus Center Postdoctoral Researcher in the School of Electrical Engineering and Computer Science at KTH, Royal Institute of Technology in Stockholm, Sweden. In 2018 he obtains a Juan de la Cierva Incorporation grant in Spain to work at ICMAT in the Geometric Mechanics and Control group. In 2019 he obtained a Junior Leader Incoming grant from Fundación La Caixa to develop control algorithms for the cooperation and coordination of multiple unmanned aerial vehicles. In 2021 he joins the Fields and Service Robotic Group at CAR.
\end{IEEEbiography}

\end{document}